\documentclass[10pt,a4paper,twocolumn]{article}
\usepackage[utf8]{inputenc}
\usepackage[T1]{fontenc}
\usepackage[top=0.9in,bottom=1.0in,left=0.7in,right=0.7in,footskip=0.45in,columnsep=0.3in]{geometry}
\usepackage{mathptmx}
\usepackage[scaled=0.92]{helvet}
\usepackage{courier}
\usepackage{graphicx}
\usepackage{booktabs}
\usepackage{microtype}
\usepackage{amsmath}
\usepackage{amssymb}
\usepackage{parskip}
\usepackage{xcolor}
\usepackage{titling}
\usepackage{enumitem}
\usepackage[section]{placeins}
\usepackage{float}
\usepackage{caption}
\newcommand{\paretoglobalimg}{\includegraphics[width=\textwidth,height=0.42\textheight,keepaspectratio]{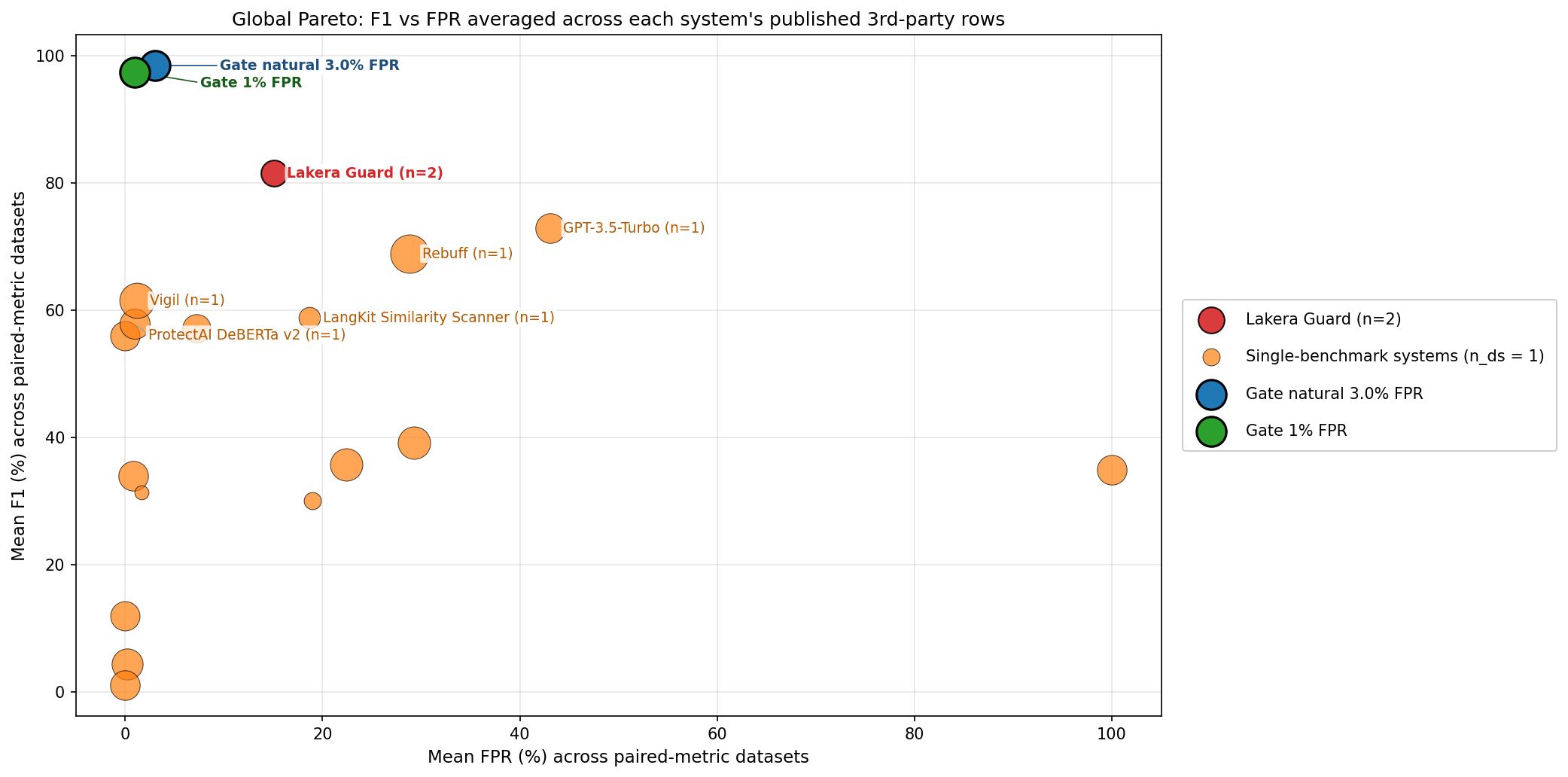}}

\usepackage{titlesec}
\usepackage{flushend}
\usepackage{multicol}
\usepackage{tabularx}
\usepackage{ragged2e}
\usepackage{needspace}

\usepackage{hyperref}
\hypersetup{
  colorlinks=true,
  linkcolor=blue!55!black,
  urlcolor=blue!55!black,
  citecolor=blue!55!black,
  pdftitle={Gate AI: LLM Security Benchmark Evaluation Methodology \& Results},
  pdfauthor={Ryle Goehausen, Marcus Sousa},
}

\titleformat{\section}{\large\bfseries}{\thesection}{0.6em}{}
\titleformat{\subsection}{\normalsize\bfseries}{\thesubsection}{0.5em}{}
\titlespacing*{\paragraph}{0pt}{0.6em}{0.25em}
\titlespacing*{\subsection}{0pt}{0.8em}{0.3em}

\pretitle{\noindent\hrulefill\par\vspace{0.5em}\begin{center}\Large\bfseries\scshape}
\posttitle{\par\vspace{0.3em}\normalfont\normalsize\textsc{A Preprint}\end{center}\vspace{0.4em}}
\preauthor{\begin{center}\normalsize}
\postauthor{\end{center}}
\predate{\vspace{1.2em}\begin{center}\small}
\postdate{\end{center}\vspace{0.3em}\noindent\hrulefill\par\vspace{0.4em}}

\renewcommand*{\thefootnote}{\fnsymbol{footnote}}
\title{Gate AI: LLM Security Benchmark Evaluation Methodology \& Results\protect\footnotemark[1]}
\author{%
  \begin{tabular}{c@{\hspace{3em}}c}
    Ryle Goehausen & Marcus Sousa \\
    \footnotesize\href{mailto:ryle@constellationnetwork.io}{\texttt{ryle@constellationnetwork.io}} &
    \footnotesize\href{mailto:marcus@constellationnetwork.io}{\texttt{marcus@constellationnetwork.io}}
  \end{tabular}%
}
\date{2026-05-27}

\begin{document}
\twocolumn[\begin{@twocolumnfalse}
\maketitle
\vspace{0.6em}

\begingroup
\leftskip1.8in \rightskip1.8in
\footnotesize
\begin{abstract}
\noindent
Published evaluations of prompt-injection and jailbreak detectors
for Large Language Models often suffer from two systematic
weaknesses: per-dataset threshold tuning and undisclosed
operating points. We describe an evaluation harness that
addresses both. The detector under evaluation is scored across
16 public benchmarks (12,111 samples) using
5-fold cross-validation. StratifiedKFold (by row) is the headline
pass; a parallel StratifiedGroupKFold pass over a composite key
(parent-prompt id plus MinHash + LSH near-duplicate clusters at
Jaccard $\gtrsim 0.8$) runs alongside it as a leakage-premium
diagnostic.
A single global operating point is selected on the held-out
folds (max F1 subject to FPR $\leq 1\%$) and applied
uniformly to every dataset, so per-dataset results reflect one
threshold rather than per-benchmark optimisation. Generalisation
is examined through a battery of diagnostics (leave-one-dataset-out
cross-validation, a random-label control, adversarial validation,
permutation feature importance, length-bias correlation,
classifier-head agreement, cross-source near-duplicate detection,
threshold transferability, train-vs-OOF agreement, and a
paraphrase-invariance probe), most with a quantitative pass
threshold and the remainder with a stated failure mode. For every external comparison, the detector's
threshold is re-tuned to the competitor's published false-positive
rate so head-to-head values are evaluated at matched operating
points.
\end{abstract}

\vspace{1.2em}
\textperiodcentered{} Benchmark
\endgroup

\vspace{2.2em}
\end{@twocolumnfalse}]

\footnotetext[1]{Working preprint. Subsequent versions may update benchmark numbers, dataset composition, or methodology as the framework evolves; refer to the most recent version for current figures.}
\renewcommand*{\thefootnote}{\arabic{footnote}}
\setcounter{footnote}{0}

\section{Introduction}

A representative deployment for the systems examined here is an
agentic assistant with read access to a user's email inbox and
write access to outgoing mail. Once attacker-supplied content
reaches the assistant (a phishing email, a hostile attachment,
a poisoned calendar invite), a single line of natural-language
instruction inside that content can cause the model to draft and
dispatch mail on the attacker's behalf, exfiltrate prior thread
contents, or invoke any other tool the assistant has access to.
The same threat surface appears in retrieval-augmented chat,
browser-automation agents, and any application where untrusted
data is mixed into an LLM's prompt. Defensive systems that filter
these attacks have proliferated, but their published evaluations
are uneven: ad-hoc datasets, undisclosed thresholds, per-dataset
tuning, and inconsistent definitions of what constitutes a positive
label make cross-system comparison difficult.

This report describes the evaluation harness used to benchmark the
detector under test against published competitor numbers. The work
is intentionally scoped to non-proprietary testing methodology: how
the trace is assembled, how the cross-validation prevents
sibling-chunk leakage, how a single global operating point is
selected and applied uniformly, how per-dataset matched-FPR
comparisons re-tune the threshold to neutralise FPR mismatches, and
how leave-one-dataset-out and random-label diagnostics stress-test
generalisation. The detector itself is treated as a black box. The
remainder of the paper is organised as follows. Section~2 describes
the testing methodology in full: trace assembly, leakage-resistant
cross-validation, inner-validation and threshold selection,
per-chunk-to-per-prompt aggregation, micro vs macro choice,
operating-point selection, the matched-FPR comparison protocol,
bootstrap confidence intervals, calibration, and the generalisation
diagnostics that run on every release alongside the empirical
results that ground them (per-fold threshold stability,
leave-one-dataset-out, random-label, calibration tables), plus
notes on determinism, limitations, and pretraining contamination.
Section~3 describes the dataset corpus and its attack-family
composition. Section~4 reports aggregate results, per-dataset
comparisons, and a head-to-head with the most-published commercial
competitor. Section~5 reports end-to-end latency. An appendix
glossary defines every metric and term used in the paper.

\section{Methodology}

\subsection{Datasets and trace assembly}

The evaluation trace combines 16 public benchmarks
covering balanced collections, all-attack adversarial corpora, and
all-benign over-defense benchmarks. Each upstream dataset is loaded
through a single source-tagged loader and the resulting samples are
unioned into one trace. Trace identity is content-hashed over the
loader file contents and a per-loader sample cap, so re-runs against
the same loader version and cap produce bit-identical traces; this
makes cache reuse safe and reproducibility verifiable. Per-dataset
descriptions and citations appear in Section~3.

\subsection{Cross-validation}

Out-of-fold predictions come from $K=5$ cross-validation. The trace
$\mathcal{D} = \{(x_i, y_i)\}_{i=1}^{N}$ is partitioned into
disjoint folds $\mathcal{D} = \bigsqcup_{k=1}^{K} \mathcal{D}_k$.
Two splitters run in parallel:

\begin{itemize}[leftmargin=1.5em,itemsep=0.18em]
  \item \textbf{StratifiedKFold} preserving the label marginal
        $P(y \mid \mathcal{D}_k) \approx P(y \mid \mathcal{D})$
        for every $k$. This is the headline pass; the trained
        model and headline F1 / FPR come from it.
  \item \textbf{StratifiedGroupKFold (diagnostic)} with a
        \emph{composite} group-key $g(\cdot)$ that is the union
        (under union-find) of two membership relations:
        \emph{primary} key: parent-prompt index when the upstream
        pipeline emits chunked rows, else the row id; and
        \emph{near-duplicate} key: MinHash + LSH
        clusters on 5-character shingles, calibrated so rows with
        Jaccard similarity $\gtrsim 0.8$ collide. Rows that share
        either key collapse into the same group transitively. The
        splitter places every member of a group in the same fold
        (no within-group leakage) while keeping the per-fold label
        marginal close to $P(y \mid \mathcal{D})$.
\end{itemize}

The two passes use the same model, same features, same
inner-valid early-stopping / threshold methodology; only the
fold-assignment policy differs. The gap
$\Delta F_1 = F_1^{\text{strat}} - F_1^{\text{sgk}}$
estimates the residual leakage premium that exact-text-identity
folds would have captured but the composite-grouped folds reject.
The earlier paper revision used plain \textbf{GroupKFold} for
this diagnostic; that conflated leakage with class-marginal
imbalance across folds (GroupKFold does not stratify by label) and
turned $\Delta F_1$ into a loose upper bound. Switching the
diagnostic to StratifiedGroupKFold removes the class-marginal
term, so $\Delta F_1$ on a clean trace tracks leakage alone.

On a per-prompt trace with no chunked rows and no near-duplicate
clusters, the composite key degenerates to row-identity, the
diagnostic partition matches a random stratified KFold, and
$\Delta F_1$ is uninformative in that regime; we say so
explicitly to avoid claiming protection the diagnostic cannot provide.

\begin{figure}[t]
\centering
\includegraphics[width=\linewidth]{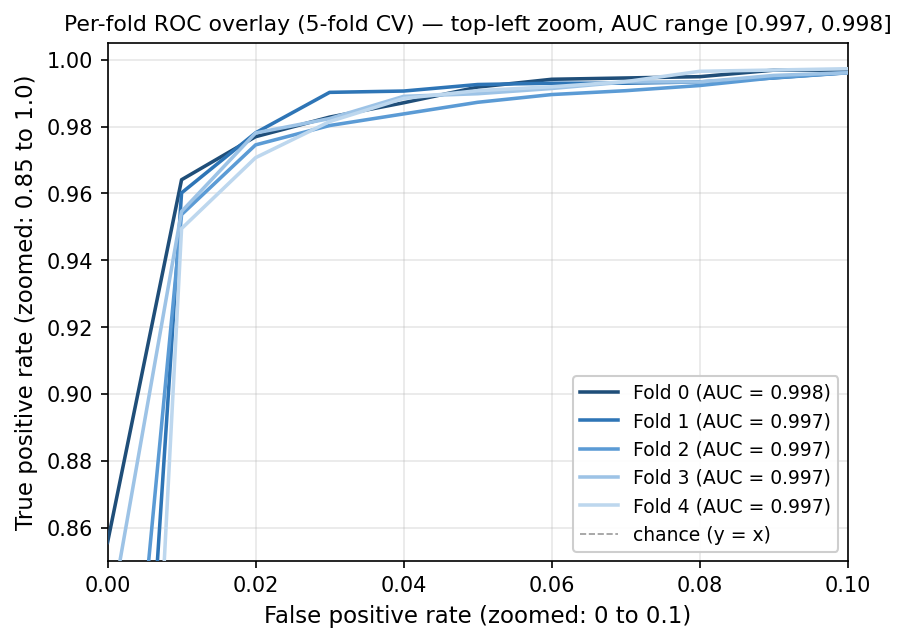}
\caption{Per-fold ROC overlay for the headline 5-fold cross-validation. Each curve is one held-out fold's score distribution against ground truth; tight clustering around a common envelope is the leak-free signal that fold-to-fold performance is stable.}
\label{fig:per-fold-roc}
\end{figure}

\subsection{Inner validation, early stopping, and threshold selection}

Each outer fold's training set
$\mathcal{D}_k^{\text{train}} = \mathcal{D} \setminus \mathcal{D}_k$
is split into stratified inner-train and inner-valid subsets in an
85 / 15 ratio. The inner-valid slice is the early-stopping
monitor (max 800 iterations, patience 40, log-loss on the
inner-valid slice as the stopping criterion) and the
operating-threshold selector. The per-fold threshold is
\[
\theta_k^* = \arg\max_{\theta \in \Theta_k}\,
  F_1\!\left(\hat{y}_\theta,\, y \,\middle|\,
              \mathcal{D}_k^{\text{inner-valid}}\right),
\]
where $\Theta_k$ is the union of all observed inner-valid
probabilities and a fallback grid
$\{0.3, 0.4, 0.5, 0.6, 0.7, 0.8, 0.9\}$. The chosen $\theta_k^*$
is applied to the held-out test fold $\mathcal{D}_k$ for hard
labels. The canonical operating point is
$\tilde\theta = \operatorname{median}_k \theta_k^*$. The test fold
is never an evaluation target during training.

\paragraph{Per-fold threshold stability (empirical).}

The 5 inner-valid-picked thresholds for the headline StratifiedKFold pass:

{\footnotesize\setlength{\tabcolsep}{4pt}\begin{tabular*}{\linewidth}{@{\extracolsep{\fill}}rrrr@{}}
\toprule
Fold $k$ & $\theta_k^*$ & $F_1^{(k)}$ & best-iter \\
\midrule
0 & 0.650 & 98.61\% & 235 \\
1 & 0.200 & 98.68\% & 208 \\
2 & 0.500 & 98.48\% & 215 \\
3 & 0.400 & 98.64\% & 294 \\
4 & 0.550 & 98.60\% & 232 \\
\bottomrule
\end{tabular*}}

Mean $\theta_k^* = 0.460$, median $\tilde\theta = 0.500$ (the canonical operating point defined in \S2.3), $\sigma = 0.153$, range $[0.200, 0.650]$. Tight clustering around $\tilde\theta$ is the leak-free signal: each fold independently arrived at a similar threshold from its own held-out inner-valid slice, without ever seeing its test fold.

\paragraph{Adversarial validation.}
Train an auxiliary classifier $g_\phi: x \mapsto [0,1]$ to predict
whether each row came from train or test:
$\phi^* = \arg\min_\phi \tfrac{1}{N} \sum_i
   \bigl( g_\phi(x_i) - \mathbb{1}[i \in \text{test}] \bigr)^2$.
A well-balanced split yields $\mathrm{AUC}(g_{\phi^*}) \approx 0.5$.
Any meaningful lift means the OOF metric is conflating distribution
shift with detection signal.

\IfFileExists{plots/adversarial-validation.png}{%
\begin{figure}[H]
  \centering
  \includegraphics[width=\linewidth]{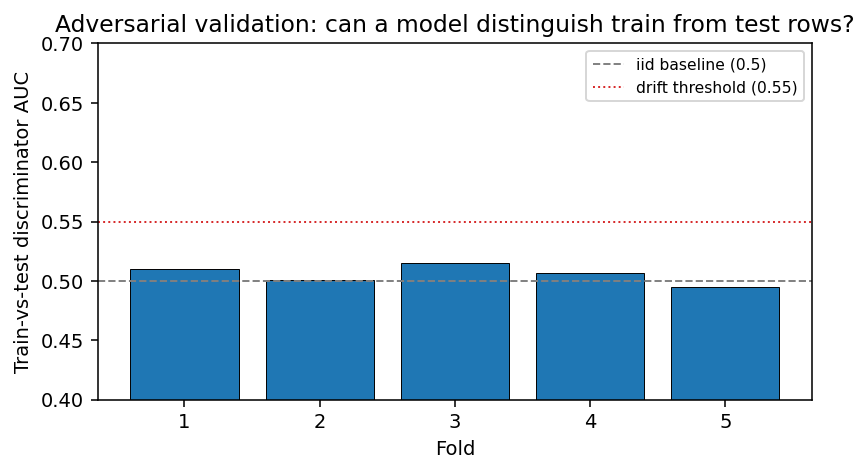}
  \caption{Adversarial-validation AUC per fold (target $\approx 0.5$).}
\end{figure}
}{}

\paragraph{Train-vs-OOF agreement.} Score the per-fold model on
its own training rows and compare to the OOF score on the same
fold. Define the gap
$\Delta^{(k)} = F_1^{\text{train}, k} - F_1^{\text{OOF}, k}$.
A small mean $\bar\Delta$ over folds confirms the OOF metric is not
under-reporting due to a pipeline-state mismatch between train and
score paths; a large $\bar\Delta$ signals overfitting.

\IfFileExists{plots/train-vs-oof-agreement.png}{%
\begin{figure}[H]
  \centering
  \includegraphics[width=\linewidth]{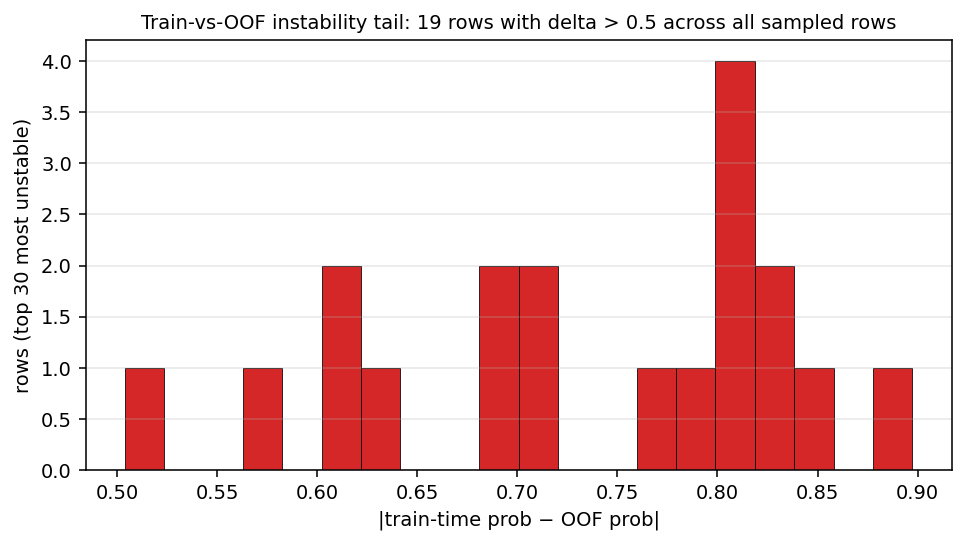}
  \caption{Per-fold train vs OOF F1.}
\end{figure}
}{}

\subsection{Per-chunk to per-prompt aggregation}

When the pipeline chunks long inputs, training operates per chunk and
aggregates at metric time. Let chunks of parent prompt $p$ be indexed
by $c \in C(p)$. Continuous probabilities are max-pooled, and the
hard label is the single global threshold $\tilde\theta = \operatorname{median}_k \theta_k^*$
(from \S2.3) applied to the max-pooled probability:
\[
\hat{p}_p = \max_{c \in C(p)} \hat{p}_{p,c},
\qquad
\hat{y}_p = \mathbb{1}\!\left[\hat{p}_p \ge \tilde\theta\right].
\]
Sample-level F1 / FPR / precision / recall come from this hard
label; AUC comes from $\hat p_p$ (threshold-free).

\paragraph{Per-fold vs aggregate operating point.} The per-fold
threshold table in \S2.3 reports the per-fold inner-valid-chosen
$\theta_k^*$, and the per-fold F1 in that table is evaluated at
$\theta_k^*$ on fold $k$'s held-out rows (per-fold operating point).
The aggregate F1/FPR reported in \S3 use the single
$\tilde\theta$ applied to every row's max-pooled probability (global
operating point). Per-fold F1 and aggregate F1 are therefore
\emph{not directly comparable} on chunked traces where sibling
chunks of one parent can land on different folds; the per-fold
table is for stability diagnostics, the aggregate is the headline
number that ships with one threshold. On the trace evaluated in
this paper every prompt fits in a single chunk so this aggregation
reduces to identity ($|C(p)|=1$ for all $p$) and the two operating
points coincide; the formulae are reported because the same
pipeline scores chunked production traffic where $|C(p)| > 1$.

\IfFileExists{plots/f1-vs-fpr-sweep.png}{%
\begin{figure}[H]
  \centering
  \includegraphics[width=\linewidth]{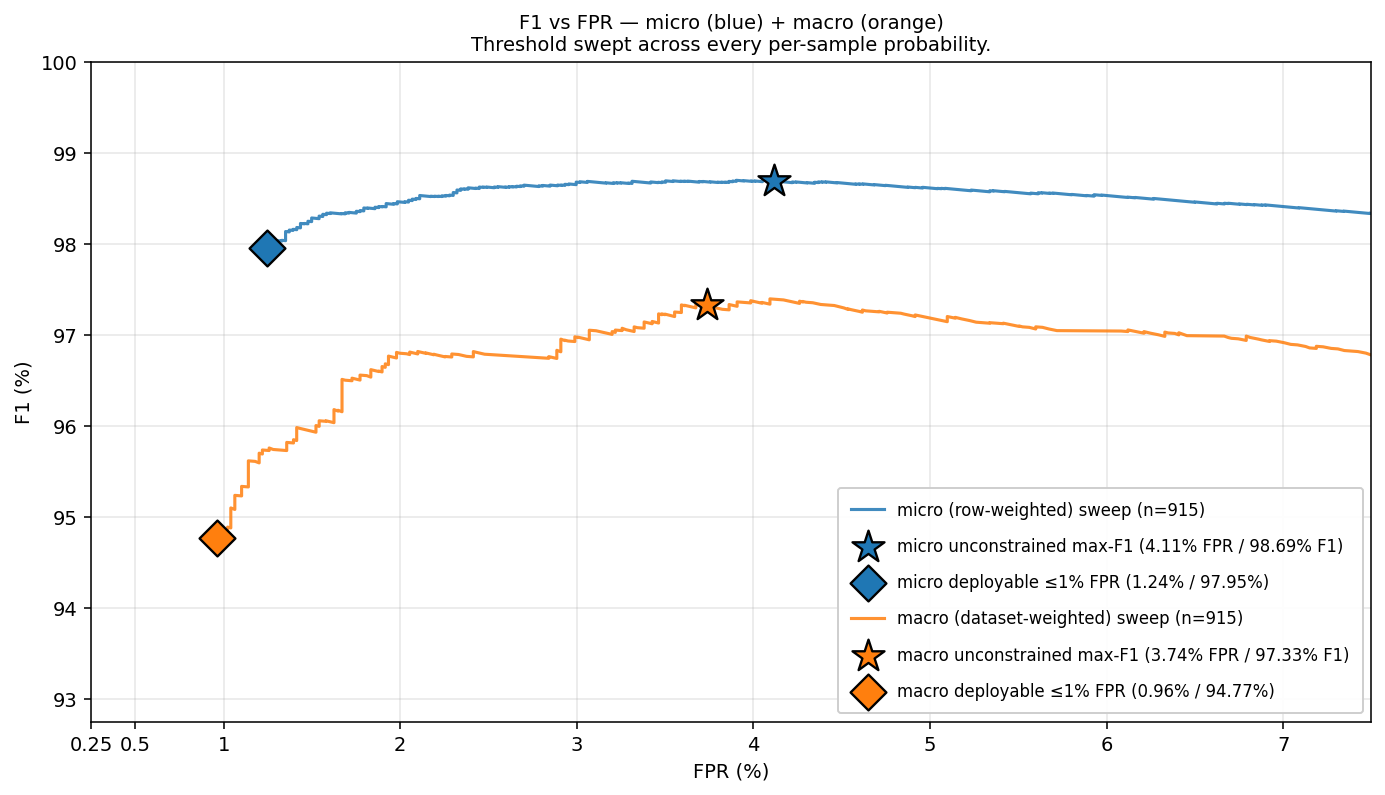}
  \caption{Cascade F1 vs FPR sweep, micro (blue) and macro (orange)
    on the same axes. Each curve sweeps the global threshold $\theta$
    across the full OOF range. The micro curve pools the per-source
    confusion matrices and is therefore dominated by the bigger
    sources; the macro curve takes the unweighted per-source F1 mean
    and gives every source the same weight. A meaningful gap between
    the two curves at the same FPR is the visual signature of
    per-source skew: when macro sits below micro the model is leaning
    on a few large sources at the expense of smaller ones; when
    macro sits above micro the bigger sources are dragging the pooled
    metric down. The headline operating point (FPR\,$\le 1\%$) and
    natural threshold are marked on both curves.}
\end{figure}
}{}

\subsection{Aggregation: micro vs macro}

Let $\mathcal{S}$ be the set of source datasets. Micro-aggregates
pool the confusion matrix across all rows, then derive metrics:
\[
F_1^{\text{micro}}
  = \frac{2 \cdot \mathrm{TP}_{\text{pool}}}
         {2\,\mathrm{TP}_{\text{pool}} +
           \mathrm{FP}_{\text{pool}} +
           \mathrm{FN}_{\text{pool}}}.
\]
Macro-aggregates take the unweighted mean across sources, skipping
single-class slices where the metric is undefined:
\[
F_1^{\text{macro}}
  = \frac{1}{\lvert \mathcal{S}_{\text{def}} \rvert}
    \sum_{s \in \mathcal{S}_{\text{def}}} F_1^{(s)}.
\]
Per-source positive rates vary widely (0\% on benign-only sets,
100\% on all-attack sets, balanced mixes elsewhere), so micro is
dominated by larger sources while macro weights all sources equally.
Both views are reported; large gaps indicate per-source skew worth
examining.

\paragraph{Prevalence ratio per source.} Per-source positive rate
$p_s = \mathbb{E}_{x \in s}[y]$ in the train mix vs the eval
mix:
$\rho_s = p_s^{\text{eval}} / p_s^{\text{train}}$.
A $\rho_s$ far from $1$ on any source signals the evaluation has
been re-balanced in a way that flatters or punishes the macro
number; we report $\rho_s$ alongside per-source results.

\IfFileExists{plots/prevalence-comparison.png}{%
\begin{figure}[H]
  \centering
  \includegraphics[width=\linewidth]{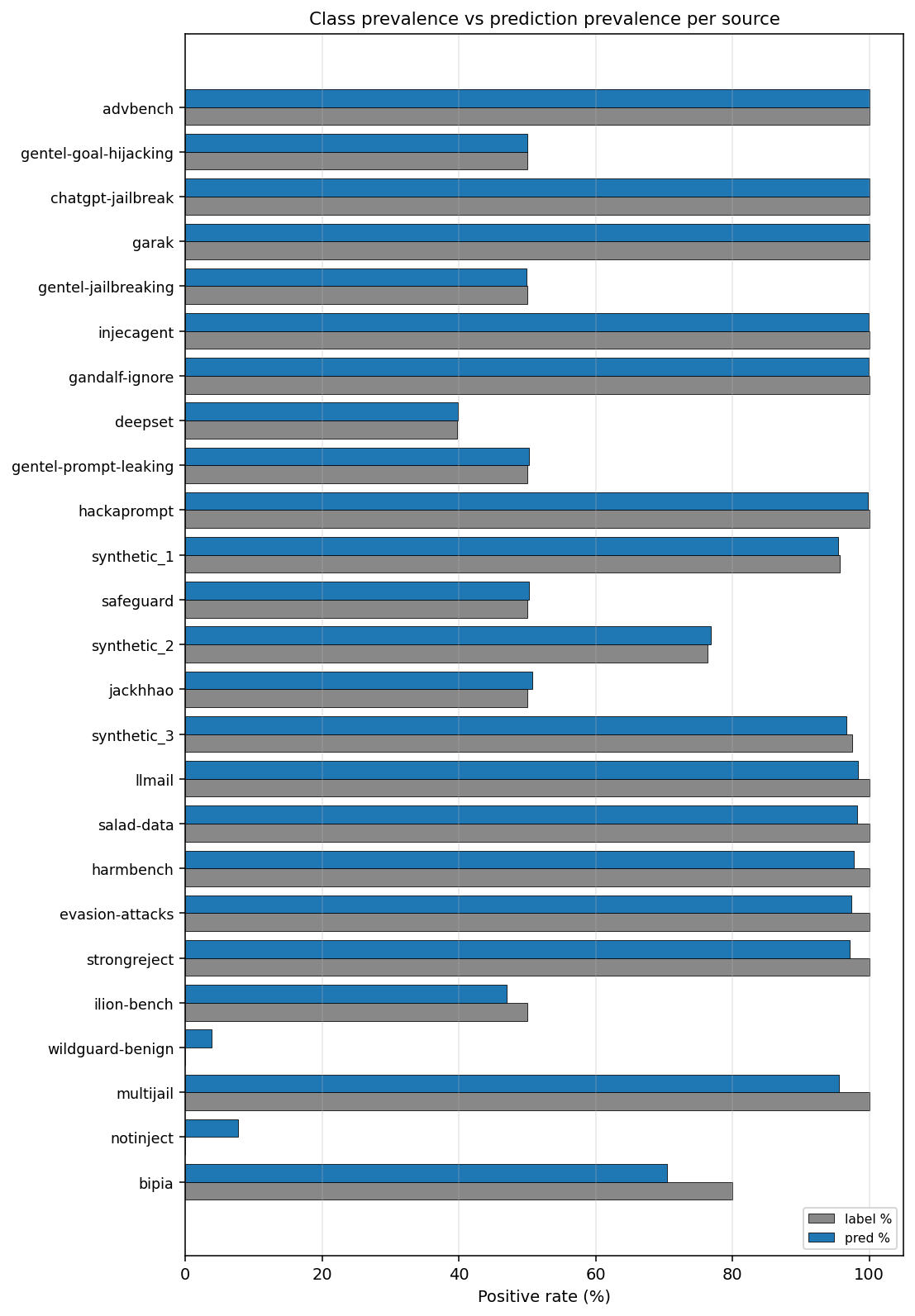}
  \caption{Train vs eval positive-rate ratio $\rho_s$ per source.}
\end{figure}
}{}

\subsection{Operating-point selection}

A single global operating point applies uniformly to every dataset.
There is no per-dataset threshold tuning and no per-dataset
hand-tuned decision rule. The operating point is the threshold
combination that solves
\[
\theta^{\text{op}}
  = \arg\max_{\theta} F_1(\hat{y}_\theta, y)
  \quad \text{subject to} \quad
  \mathrm{FPR}(\hat{y}_\theta, y) \le \tau,
\]
solved by exhaustive search over a coarse-but-well-spread candidate
grid on out-of-fold predictions, with $\tau$ the target FPR
(this run: $\tau = 1\%$). $\tau$ is a deployment
choice, not a theoretical constant: $1\%$ approximates the
operator-reported false-alarm rate above which downstream users
report alert fatigue and start ignoring or bypassing the
detector. The same $\theta^{\text{op}}$ then scores every
dataset; per-dataset values reflect what one global threshold
yields, not the best achievable in isolation. The threshold is
reported alongside the document so the operating point is
auditable.

\paragraph{Threshold transferability.} For each source $s$ we
re-pick the threshold $\theta_s^*$ that achieves the same target FPR
on $s$-only rows. We report the spread
$\sigma_\theta = \mathrm{stddev}(\{\theta_s^* - \theta^{\text{op}}\}_{s \in \mathcal{S}})$.
A tight spread is the signature of a transferable operating point;
a wide spread means a single global $\theta^{\text{op}}$ is
mis-calibrated for some sources.

\IfFileExists{plots/threshold-transferability.png}{%
\begin{figure}[H]
  \centering
  \includegraphics[width=\linewidth]{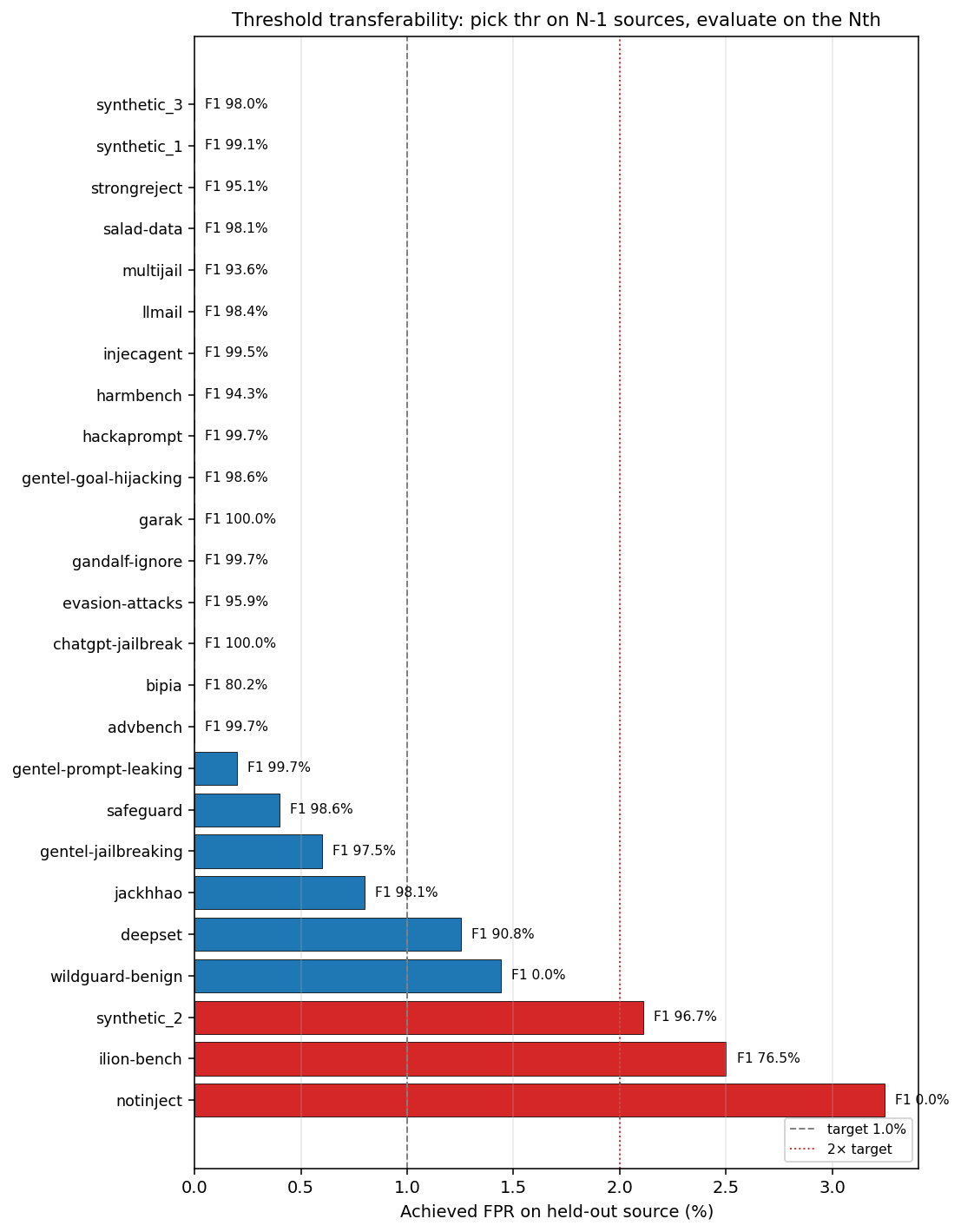}
  \caption{Per-source $\theta_s^*$ at matched FPR vs the global
    $\theta^{\text{op}}$. Each dot is a source.}
\end{figure}
}{}

\subsection{Matched-FPR per-dataset comparisons}

Comparing detectors at the same global threshold can mislead when
competitors publish their numbers at very different operating
points. To remove that confound, every per-dataset competitor entry
that publishes both a primary metric and an FPR is also surfaced
with our value re-evaluated at the threshold whose FPR matches
theirs on that dataset. The result is a like-for-like comparison at
the same FPR per row. Two surrogate paths apply on a subset of
rows so the reader can tell them apart from a true match.
\textbf{(a)} On datasets whose primary metric is itself FPR or
over-defense-accuracy (notinject, wildguard-benign), matching FPR
collapses the comparison; our value at the global headline
threshold is shown instead and tagged $^\dagger$ in both the
per-dataset table and the Lakera overview plot.
\textbf{(b)} When a competitor publishes only the primary metric
without an FPR (which Lakera does on the three gentel-bench
attack-type splits), our threshold is matched to a 1\% FPR
surrogate rather than the competitor's unknown operating point;
those rows are tagged $^\ddagger$.

\paragraph{Threshold-selection caveat.} Picking
$\theta^{\text{matched}}_s$ to satisfy the competitor's
published FPR uses the same per-source rows that the matched-FPR
F1 is then scored on; in the threshold-on-test sense of
Cawley \& Talbot 2010~\cite{c22} this is double-dipping, and the point-estimate F1
slightly overstates what a fixed pre-registered threshold would
have produced. The matched-$n$ bootstrap CI surfaced in the
companion table mitigates this by resampling rows; rows whose
draws also re-pick $\theta$ widen the CI to include the
threshold-selection uncertainty, so when the competitor's point
estimate falls inside the CI we mark the row as ``within noise''
rather than claim a win. The headline global-$\theta^{\text{op}}$
F1 in Section~3 is not affected; it is reported at a single threshold
picked from inner-valid data without seeing the per-source rows
it is scored on.

\subsection{Generalisation diagnostics}

We stress-test the headline number with a small set of integrity
checks that each carry a precise pass criterion. Below: name,
mechanism, math, and the actual diagnostic plot inline. Each
check runs as part of every evaluation release.

\paragraph{Leave-one-dataset-out (LODO).} Hold out one source
$s$ from training and evaluate Gate only on rows of $s$. For each
held-out source we compute
$F_1^{(\text{LODO}, s)} = F_1(\hat{y}_{-s}, y_s)$
where $\hat{y}_{-s}$ is the prediction from a model trained on
$\mathcal{D} \setminus \{s\}$. The macro mean
$\bar F_1^{\text{LODO}} = \frac{1}{|\mathcal{S}|}
\sum_{s \in \mathcal{S}} F_1^{(\text{LODO}, s)}$
bounds how much performance drops when a previously unseen
distribution arrives.

\begin{figure}[H]
\centering
\includegraphics[width=\linewidth]{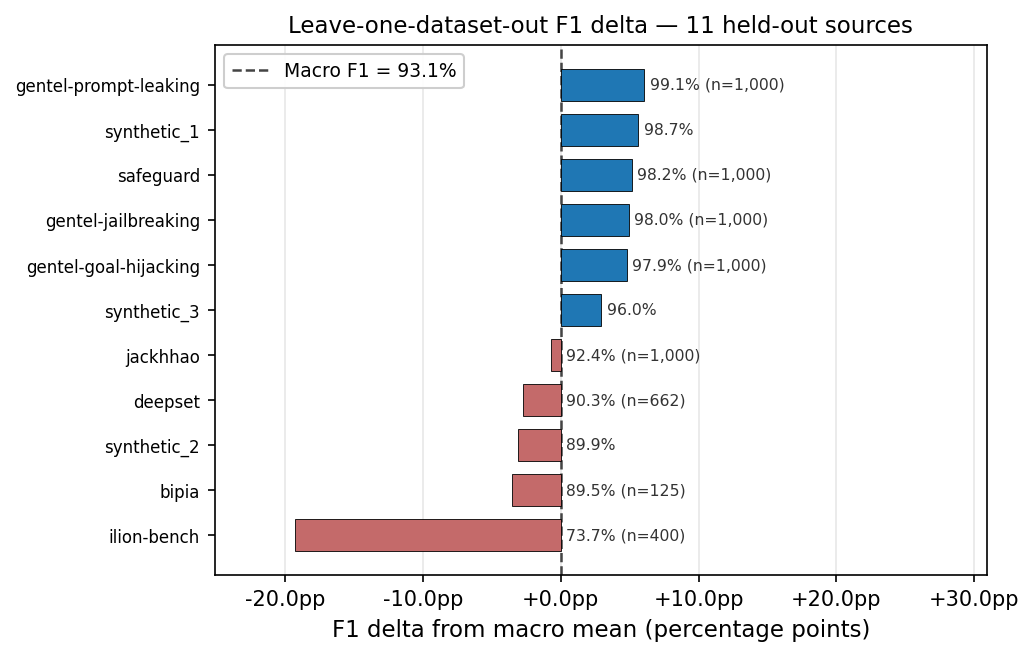}
\caption{Leave-one-dataset-out F1 delta from the macro mean per held-out source. Bar = (per-source F1) $-$ (macro F1) in percentage points; macro mean at 0. Blue $\ge$ macro, red below. Held-out F1\% and sample count $n$ annotated at right of each row.}
\label{fig:lodo-forest}
\end{figure}

\texttt{ilion-bench} sits substantially below the macro mean. Its prompt distribution is remarkably different from the rest of the public benchmark cohort because of its role-playing structure.

\paragraph{Random-label control.} Shuffle the labels
$\tilde y = \pi(y)$ where $\pi$ is a uniform permutation of
$\{1, \dots, N\}$ and score the model's existing hard predictions
$\hat y$ against the shuffled targets. Under independence, the
expected F1 is
\[
F_1^{\text{chance}}
  = \frac{2\, p_y\, p_{\hat y}}{p_y + p_{\hat y}},
\]
where $p_y = \mathbb{E}[y]$ is the label prevalence and
$p_{\hat y} = \mathbb{E}[\hat y]$ is the predicted-positive rate.
Anything notably above $F_1^{\text{chance}}$ is row-identity
leakage. (A retrain-from-scratch variant of this check, in which the
full pipeline is refit on shuffled labels, is also supported but is
not part of the cheap CI diagnostic.)

\IfFileExists{plots/methodology-random-label.png}{%
\begin{figure}[H]
  \centering
  \includegraphics[width=\linewidth]{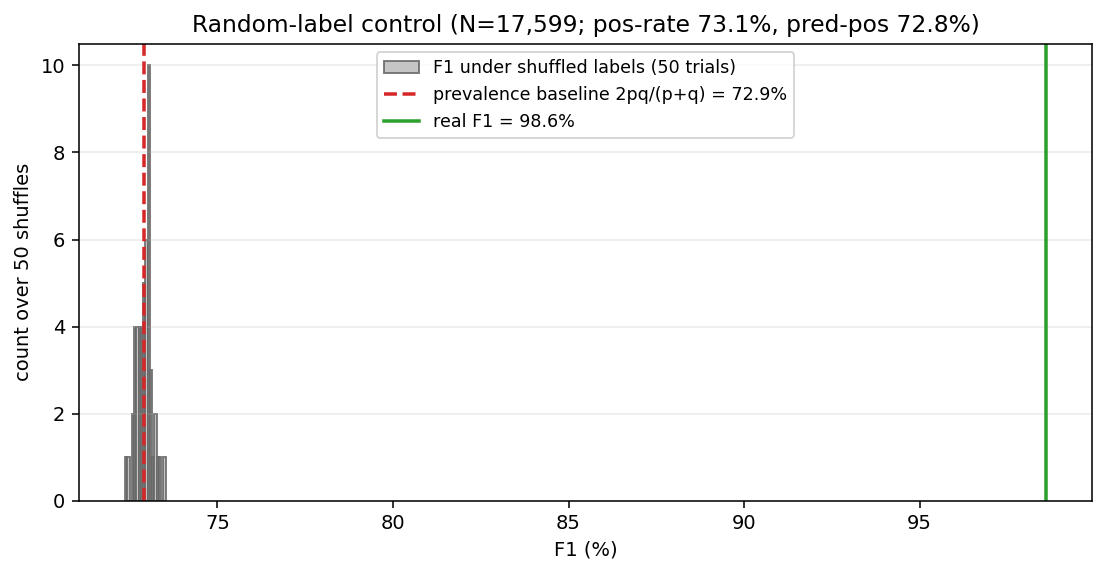}
  \caption{Random-label control: predicted hard labels scored
    against shuffled targets collapse to the chance F1 baseline
    $F_1^{\text{chance}} = 2 p_y p_{\hat y} / (p_y + p_{\hat y})$,
    confirming no row-identity leakage.}
\end{figure}
}{}

\paragraph{Random-label result (empirical).} Under shuffled labels: AUC = \textbf{0.5146} $\cdot$ F1 = \textbf{84.44\%} $\cdot$ always-positive baseline $2p_y/(1+p_y) = 84.44\%$ (upper bound on chance F1; $p_{\hat y}$ unavailable). Both statistics sit at chance level; no exploitable signal survives label permutation, so out-of-fold predictions are not leveraging row-identity leakage.

\paragraph{Length-bias correlation.} Pearson correlation between
the input text length $|x|$ and Gate's score $\hat p$,
$\rho_{\text{len}} = \mathrm{corr}(|x|, \hat p)$,
overall and per-source. A strong $|\rho| > 0.3$ flags a shallow
detector that is effectively a length heuristic: longer prompts
have more surface area for malicious tokens to land in, so a
detector that learns to flag long inputs would inherit some of
the signal without actually understanding the attack. The
per-source breakdown matters because a global correlation near
zero can hide source-specific biases that cancel out in
aggregate. Sources where benign and attack examples differ
sharply in length are flagged separately so the per-source
$\rho$ can be inspected.

\IfFileExists{plots/length-bias.png}{%
\begin{figure}[H]
  \centering
  \includegraphics[width=\linewidth]{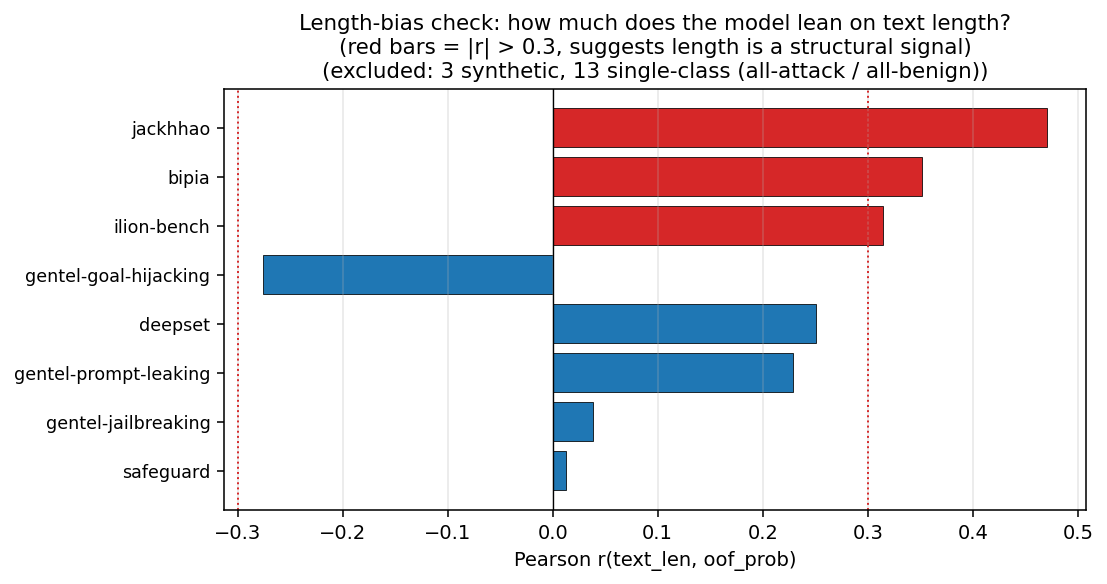}
  \caption{Length-bias Pearson correlation between input length and
    Gate score, overall and per source.}
\end{figure}
}{}

\paragraph{Permutation feature importance.} For each feature $j$,
permute its column $x_{\cdot j} \to x_{\pi(\cdot) j}$ and
re-score on OOF predictions. The importance is the drop in F1:
$\Delta F_1^{(j)} = F_1(\hat y, y) - F_1(\hat y^{\pi_j}, y)$.
Large $\Delta F_1^{(j)}$ confirms the feature is load-bearing in
the held-out regime, beyond what a training-time importance
ranking can credit, since the latter can attribute weight to
features that fail to generalise out of fold.

\IfFileExists{plots/permutation-importance.png}{%
\begin{figure}[H]
  \centering
  \includegraphics[width=\linewidth]{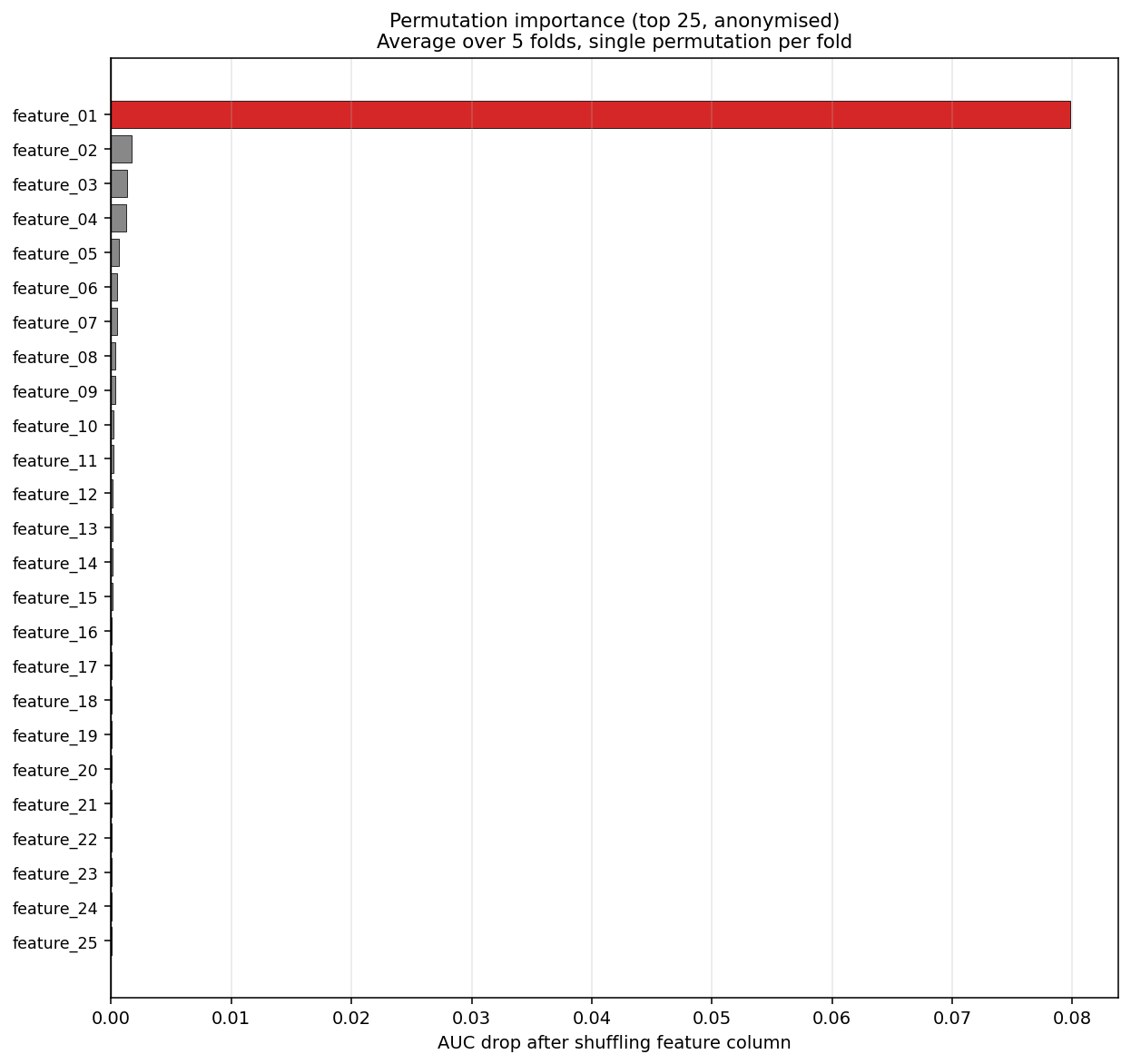}
  \caption{Top permutation-importance features by held-out F1 drop.}
\end{figure}
}{}

\paragraph{Classifier-head agreement (Cohen's \texorpdfstring{$\kappa$}{kappa}).}
Pairwise Cohen's $\kappa$ between the hard predictions of the
ensemble heads:
$\kappa = \frac{p_o - p_e}{1 - p_e}$
where $p_o$ is observed agreement and $p_e$ is chance agreement.
Low $\kappa$ values confirm the heads carry independent signal
worth combining; $\kappa \to 1$ would mean the heads are
redundant. Head names are anonymised in the rendered figure. \emph{Empirical:} pairwise $\kappa \in [0.45, 0.56]$ across 3 head pairs; the heads carry distinguishable signal.

\IfFileExists{plots/classifier-agreement.png}{%
\begin{figure}[H]
  \centering
  \includegraphics[width=\linewidth]{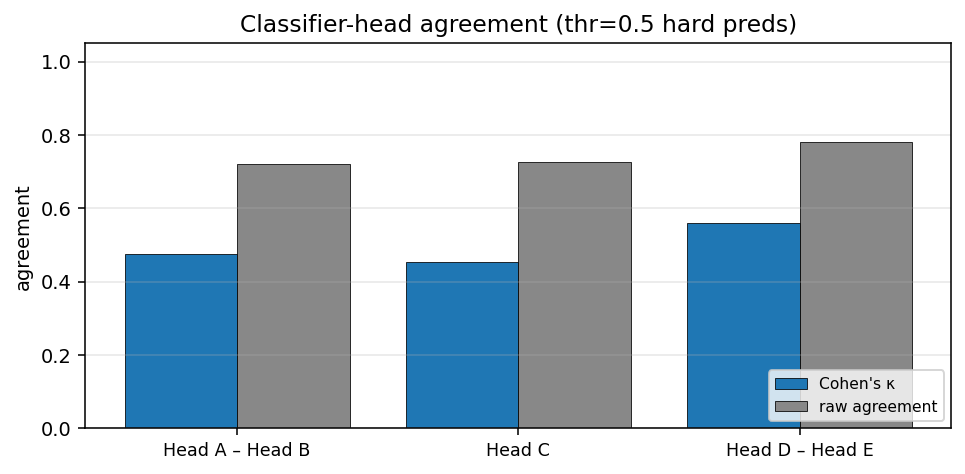}
  \caption{Pairwise Cohen's $\kappa$ between ensemble heads.}
\end{figure}
}{}

\paragraph{Other checks (summary only).} Cross-source
near-duplicate hashing catches prompts that appear in multiple
datasets under conflicting labels (the most common form of
implicit data-quality leak when stitching public benchmarks),
and surfaces them for manual relabel; determinism replay diffs
two runs with identical seed for byte-equal OOF probabilities;
a paraphrase-invariance check bounds how lexically-shallow the
final-stage signal is.

\subsection{Confidence intervals and bootstrap}

Every headline metric (F1, precision, recall, FPR, AUC) is reported
with a 95\% stratified bootstrap confidence interval. Resamples are
drawn within stratification cells
$(s, y) \in \mathcal{S} \times \{0,1\}$ (source $\times$ label),
so rare-class slices do not collapse; within each cell $(s,y)$ we
draw $n_{s,y}$ rows with replacement, so the total resample size
equals the original $N$ exactly. With $B = 10\,000$ resamples
and per-resample metric estimates
$\hat{\theta}^{*(1)}, \dots, \hat{\theta}^{*(B)}$, the
percentile interval is
\[
\mathrm{CI}_{95}(\hat\theta)
  = \bigl[\, \hat{\theta}^*_{(\lceil 0.025 B \rceil)},\
                \hat{\theta}^*_{(\lfloor 0.975 B \rfloor)} \,\bigr].
\]
Per-dataset CIs are stratified by label only. Datasets with $n <
200$ are flagged as small-n (their CI is correspondingly wide). When
a competitor's published point estimate falls inside our CI we
treat the comparison as within noise rather than claiming a win.
$B$ controls tail-stability of the CI (with $B=10\,000$ each tail
is determined by $\approx\!250$ resamples, reproducible across
seeds to better than a column-width of the CI); the CI width itself
is driven by $n$.

The full-page forest figure visualises this directly: Gate's
per-dataset point estimate is a tapered blue band (50\% CI core, 80 /
95 / 99\% tails), grouped by attack family; red diamonds mark Lakera
Guard, grey dots every other published competitor (asterisk =
self-evaluation, treated as an upper bound).

\subsection{Calibration and threshold sweeps}

Probability calibration is reported pre- and post-isotonic
regression. Brier score and Expected Calibration Error (ECE) are
computed over $M=15$ equal-width confidence bins:
\[
\mathrm{ECE} = \sum_{m=1}^{M}
  \frac{|B_m|}{N}
  \bigl| \operatorname{acc}(B_m) - \operatorname{conf}(B_m) \bigr|,
\]
\[
\mathrm{Brier} = \frac{1}{N}
  \sum_{i=1}^{N} (\hat{p}_i - y_i)^2.
\]
A reliability diagram checks whether the detector's output
probabilities are distribution-calibrated against the empirical
positive rate at each predicted-probability bucket.

\paragraph{Calibration (empirical).} Pre- and post-isotonic regression over $M=15$ equal-width bins:

{\footnotesize\setlength{\tabcolsep}{4pt}\begin{tabular*}{\linewidth}{@{\extracolsep{\fill}}lrr@{}}
\toprule
Metric & Pre-isotonic & Post-isotonic \\
\midrule
ECE & 0.0081 & 0.0016 \\
Brier & 0.0158 & 0.0153 \\
\bottomrule
\end{tabular*}}

\begin{figure}[H]
\centering
\includegraphics[width=\linewidth]{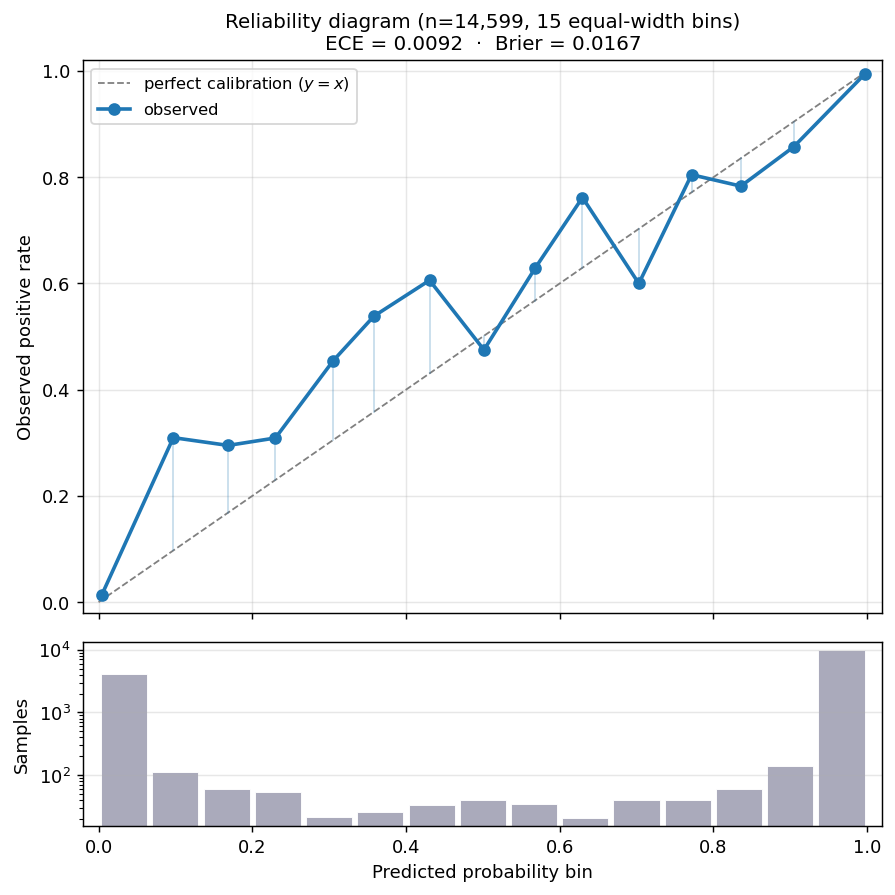}
\caption{Reliability diagram: predicted probability bin vs observed positive rate, with a per-bin sample-count histogram below.}
\end{figure}

\begin{figure*}[p]
\centering
\includegraphics[width=\linewidth,height=0.95\textheight,keepaspectratio]{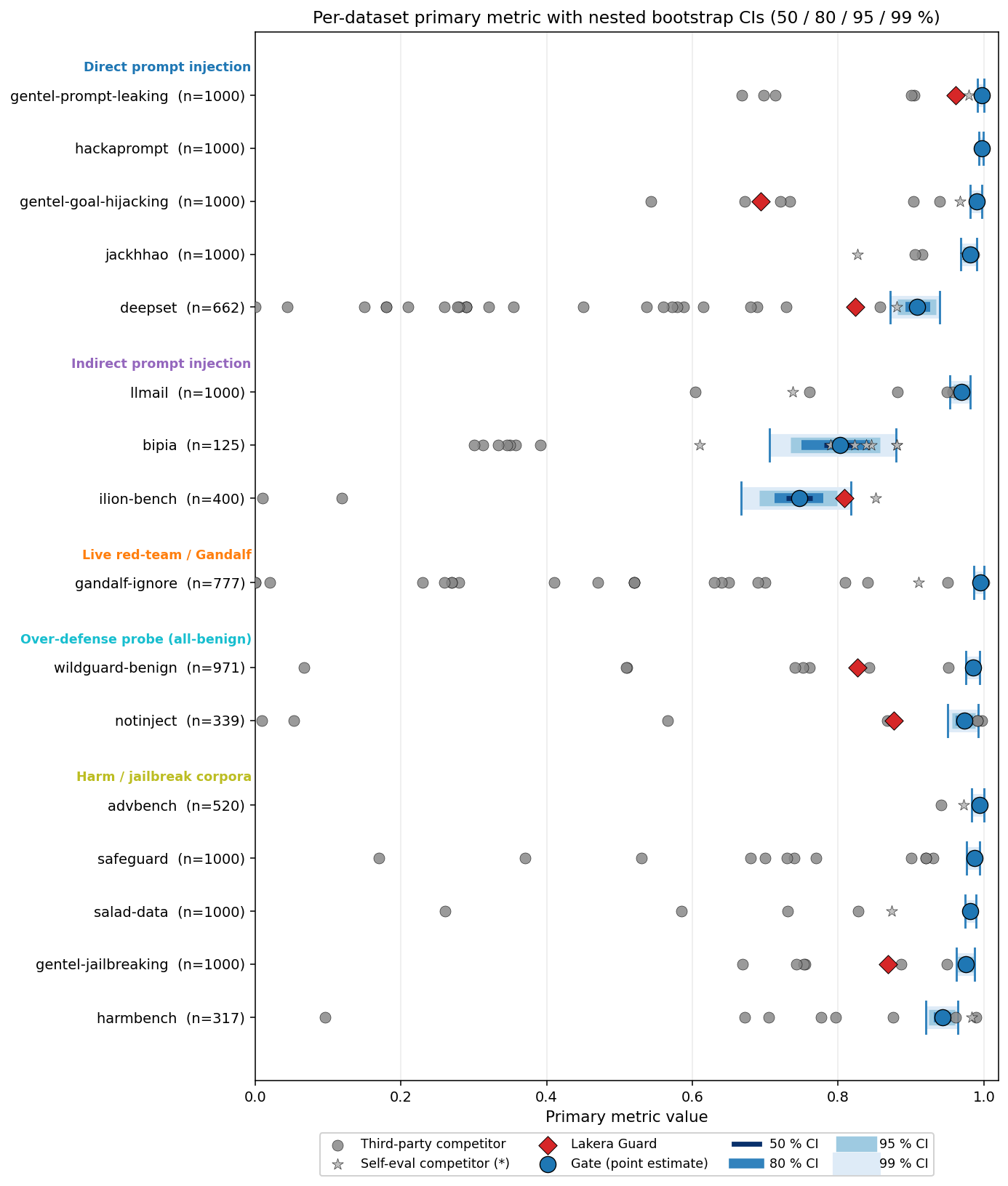}
\caption{Per-dataset primary metric with nested bootstrap CIs.}
\end{figure*}

\subsection{Comparing against external competitors}

External baselines come from third-party published numbers (papers,
model cards, dataset cards). Self-evaluations are marked with an
asterisk and treated as upper bounds because the competitor's
training split may overlap the dataset. We do not rerun frozen
published numbers on our own trace (evaluation splits, dataset
versions, and preprocessing all differ), so head-to-head deltas
are bracketed by our own bootstrap CIs, and per-dataset footnotes
record the competitor's published $n$ and the date of their
evaluation.

\subsection{Determinism and reproducibility}

Every randomised step in the evaluation pipeline is seeded from a
single \texttt{SEED} constant: the outer CV split, the inner-valid
split inside each fold, and the bootstrap-resample sequence used
for confidence intervals. Intermediate artifacts are checkpointed
to JSONL so downstream metrics replay deterministically from the
recorded artifact rather than re-running upstream computation that
may drift over time.

\subsection{Limitations}

The composite group key in Section~2 covers exact-text
duplicates (SHA-256 of normalised text) and approximate-text
duplicates (MinHash + LSH on 5-character shingles at a Jaccard
cut-off of $\approx 0.8$). MinHash + LSH was chosen for its
scalability on larger evaluation traces --- clustering cost is
near-linear in the row count, so the diagnostic remains tractable
as the corpus grows. Adversaries that mutate prompts beyond that
shingle-overlap budget (aggressive paraphrasing, language
switching, encoding-level transforms) will not be linked into the
same group and could in principle leak across folds; future
evaluations will pair MinHash with more comprehensive similarity
functions (sentence embeddings, dense-retrieval search indexes)
for the overlap calculation so semantic near-duplicates that
shingle-Jaccard misses also get caught.
Per-dataset matched-FPR comparisons use the competitor's published
per-dataset FPR as the threshold target where available, but some
sources publish only aggregate numbers, in which case the
comparison falls back to the global threshold and is tagged as
such in figures. The benchmark breadth itself is limited to
publicly available datasets with at least one published competitor
number on a metric we evaluate; the breadth grows as new academic
releases are added.

\subsection{Pretraining contamination}

Public prompt-injection datasets such as
\texttt{deepset/\allowbreak prompt-injections},
\texttt{xTRam1/\allowbreak safe-guard-\allowbreak prompt-injection},
and BIPIA are widely indexed and almost certainly appear in the
pretraining data of any modern foundation-model-derived detector. This applies to the
system under test and to every competitor built on or fine-tuned
from a publicly pretrained base, so it is not a per-system
disclaimer but a property of the field. Mitigations tracked as
follow-up work include (a) held-out evaluation on novel attack
patterns generated after foundation-model training cutoffs, and
(b) per-source comparison of test F1 against the rate at which the
source appears in publicly indexed corpora.

\section{Data}

Every dataset in this leaderboard is publicly available, with its primary metric defined a priori and its competitor citations sourced from third-party publications (papers, model cards, dataset cards) unless explicitly marked as a self-evaluation. The trace is content-hashed by its loader source set so any re-run on the same loader hash is bit-identical.

\subsection*{Trace composition by attack family}

Datasets grouped by the attack distribution they primarily stress-test. The same global operating point scores every family, without per-family tuning. 3 synthetic datasets are excluded from this headline trace and from every metric reported in Sections 4--5; they appear only in the OOF diagnostics, where they sanity-check threshold transferability. 

\begin{table*}[!tbp]\centering\footnotesize\setlength{\tabcolsep}{6pt}\begin{tabular*}{\linewidth}{@{\extracolsep{\fill}}lp{0.50\linewidth}rr@{}}
\toprule
Attack family & Datasets & Samples & \% of trace \\
\midrule
Direct prompt injection & \parbox[t]{6.2cm}{\raggedright deepset, hackaprompt, jackhhao, gentel-goal-hijacking, gentel-prompt-leaking} & 4,662 & 38.5\% \\
Indirect prompt injection & \parbox[t]{6.2cm}{\raggedright llmail, bipia, ilion-bench} & 1,525 & 12.6\% \\
Live red-team / Gandalf & \parbox[t]{6.2cm}{\raggedright gandalf-ignore} & 777 & 6.4\% \\
Over-defense probe (all-benign) & \parbox[t]{6.2cm}{\raggedright notinject, wildguard-benign} & 1,310 & 10.8\% \\
Harm / jailbreak corpora & \parbox[t]{6.2cm}{\raggedright safeguard, advbench, gentel-jailbreaking, harmbench, salad-data} & 3,837 & 31.7\% \\
\bottomrule
\end{tabular*}\end{table*}

\IfFileExists{plots/brainstorm-metric-availability.png}{%
\begin{figure}[!tbp]
\centering
  \includegraphics[width=\linewidth,height=0.22\textheight,keepaspectratio]{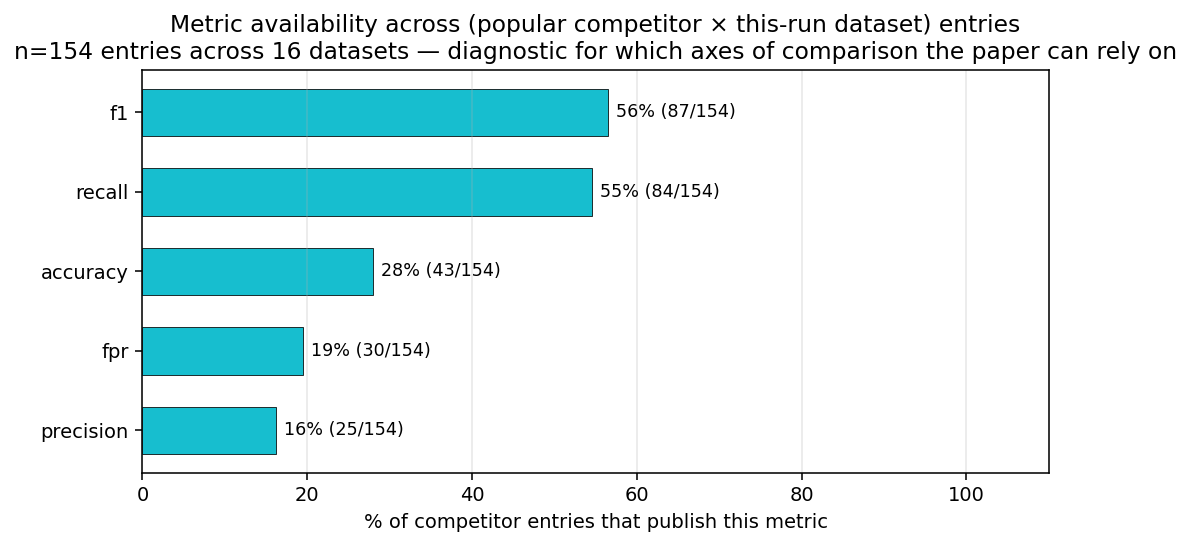}
  \caption{Metric availability across published competitor entries: F1 and recall dominate, FPR is rarest, which is why matched-FPR comparisons need per-dataset fallbacks.}
\end{figure}
}{}
\subsection*{Per-dataset detail}

\begin{table*}[!tbp]\centering\footnotesize\setlength{\tabcolsep}{4pt}\renewcommand{\arraystretch}{1.15}\begin{tabularx}{\linewidth}{l r r r >{\RaggedRight\arraybackslash}X l}
\toprule
Dataset & $n$ & \# comp. & Rank & Gate vs best comp.\ (primary metric labelled) & Gate vs Lakera (matched FPR) \\
\midrule
deepset~\cite{c5} & 662 & 27 & \#1/28 & F1\,90.8\% vs 88.0\% (+2.8, Sentinel v2*~\cite{c6}) \newline @\,matched FPR\,1.2\%: F1\,91.7\% vs 61.5\% (+30.2, Vigil~\cite{c1}) & 88.7\% vs 82.3\% (+6.4 @ FPR\,0.7\%) \\
jackhhao~\cite{c8} & 1000 & 18 & \#2/6 & F1\,98.1\% vs 98.6\% ($-$0.5, qualifire/prompt-injection-...~\cite{c6}) & n/a \\
llmail~\cite{c9} & 1000 & 6 & \#1/7 & Recall\,96.9\% vs 95.7\% (+1.2, Direct Detector~\cite{c10}) & n/a \\
notinject~\cite{c11} & 339 & 9 & \#4/10 & ODA\,97.3\% vs 99.7\% ($-$2.4, LlamaGuard-3~\cite{c11}) & 97.3\% vs 87.6\% (+9.7)$^\dagger$ \\
safeguard~\cite{c12} & 1000 & 14 & \#1/13 & F1\,98.7\% vs 93.0\% (+5.7, IBM Granite Guardian 3.2-3B~\cite{c13}) & n/a \\
advbench~\cite{c14} & 520 & 2 & \#1/3 & Recall\,99.4\% vs 97.2\% (+2.2, SmoothLLM*~\cite{c15}) & n/a \\
bipia~\cite{c16} & 125 & 23 & \#7/16 & F1\,80.2\% vs 88.0\% ($-$7.8, JavelinGuard Sharanga7*~\cite{c2}) \newline @\,matched FPR\,0.0\%: F1\,91.9\% (+3.9) & n/a \\
gandalf-ignore~\cite{c17} & 777 & 23 & \#2/24 & Recall\,99.5\% vs 100.0\% ($-$0.5, Llama-Prompt-Guard-2-86M~\cite{c13}) & n/a \\
gentel-goal-hijacking~\cite{c18} & 1000 & 8 & \#1/9 & F1\,99.0\% vs 96.7\% (+2.2, GenTel-Shield*~\cite{c18}) & 99.5\% vs 69.3\% (+30.2 @ FPR\,1.0\%)$^\ddagger$ \\
gentel-jailbreaking~\cite{c18} & 1000 & 8 & \#2/9 & F1\,97.6\% vs 97.7\% ($-$0.1, GenTel-Shield*~\cite{c18}) & 98.5\% vs 86.8\% (+11.7 @ FPR\,1.0\%)$^\ddagger$ \\
gentel-prompt-leaking~\cite{c18} & 1000 & 8 & \#1/9 & F1\,99.7\% vs 99.3\% (+0.4, WhyLabs LangKit~\cite{c18}) & 99.5\% vs 96.2\% (+3.3 @ FPR\,1.0\%)$^\ddagger$ \\
harmbench~\cite{c19} & 317 & 11 & \#5/12 & F1\,94.3\% vs 98.9\% ($-$4.6, WildGuard~\cite{c20}) & n/a \\
ilion-bench~\cite{c3} & 400 & 4 & \#3/5 & F1\,74.6\% vs 85.2\% ($-$10.5, ILION*~\cite{c3}) \newline @\,matched FPR\,7.9\%: F1\,83.6\% ($-$1.6) & 83.9\% vs 80.9\% (+3.0 @ FPR\,29.5\%) \\
salad-data~\cite{c21} & 1000 & 18 & \#1/6 & F1\,98.1\% vs 87.3\% (+10.8, MD-Judge*~\cite{c21}) & n/a \\
wildguard-benign~\cite{c20} & 971 & 28 & \#1/10 & ODA\,98.6\% vs 95.2\% (+3.4, LlamaGuard 3~\cite{c11}) & 98.6\% vs 82.6\% (+16.0)$^\dagger$ \\
\bottomrule
\end{tabularx}
\vspace{0.3em}\par
{\footnotesize \textbf{Rank column.} Gate's rank on this dataset's primary metric among all comparable systems (Gate plus every third-party-verified competitor that publishes that metric), shown as rank\,/\,total. The total counts only systems publishing the primary metric, so it can be smaller than the \# comp.\ column, which counts every competitor publishing any comparable metric (F1 / recall / precision / FPR). Computed at Gate's global headline threshold, not the matched-FPR line. \textbf{Gate vs best comp.\ column.} The main row is Gate at the global headline operating point (FPR\,$\le 1\%$) vs the best third-party-verified competitor on this dataset's \emph{primary metric} (the metric the dataset was designed to be scored on --- F1 for balanced splits, Recall for all-attack collections, ODA = $1-\text{FPR}$ for all-benign probes, FPR for over-defense corpora). Each cell leads with the named metric (F1 / Recall / Precision / Accuracy / ODA / FPR) so the comparison axis is unambiguous. The second line, where present, is the like-for-like matched-FPR comparison: Gate re-thresholded to a competitor's published FPR on this dataset, scored on the same primary metric. The matched-FPR comparator is the first competitor walked down the metric-ranked list (best first) that publishes an FPR \emph{and} for which retuning Gate to that FPR improves on Gate's value at the global headline operating point. Lakera Guard is excluded from this walk because the right-hand column already carries that comparison. When the best competitor publishes FPR and the retune improves Gate, the second line and the main row name the same comparator; otherwise the walk falls through to the highest-metric non-Lakera competitor that satisfies both conditions. When no non-Lakera competitor on the row publishes an FPR, or when every such retune would lower Gate's value, the matched-FPR line is dropped. Datasets whose primary metric is FPR or ODA (notinject, wildguard-benign) skip the second line because matching FPR is degenerate there (the matched axis equals the primary axis). \textbf{Gate vs Lakera column.} $^\dagger$ Lakera comparison shown at Gate's global headline threshold because matching FPR collapses the comparison (the dataset's primary metric is FPR / over-defense-accuracy itself). Applies to: notinject, wildguard-benign. $^\ddagger$ Lakera published the primary metric but not an FPR for this dataset, so Gate's threshold is matched to a 1\% FPR surrogate instead of Lakera's (unknown) operating point. Applies to: gentel-jailbreaking, gentel-goal-hijacking, gentel-prompt-leaking. All other Lakera-column rows are true matched-FPR comparisons (Gate's threshold re-tuned per row to hit Lakera's published FPR).}
\end{table*}

\subsection*{Coverage gaps}

Areas this report does \emph{not} cover. Customers should treat claims as scoped to the distributions actually evaluated above.

\begin{itemize}
  \item Adversarial pixel and image-payload attacks (vision-only channels). The current evaluation covers text inputs.
  \item Voice-channel jailbreaks (audio prompts transcribed inside the agent loop). Tracked as follow-up but not in this report.
  \item Languages beyond the public multilingual benchmarks (MultiJail's nine languages and SEA-SafeguardBench's eight). Coverage of low-resource and emerging-language attack patterns is a known gap.
  \item Long-context document injection beyond the chunk lengths of the public benchmarks (typical max \textasciitilde{}8 KB). Production traces with very long documents may have different distribution.
  \item Code-execution and shell-injection payloads embedded in natural-language prompts (e.g. subprocess wrappers, Python eval bait). The current trace contains a handful of such examples through harmbench and salad-data, but no dedicated benchmark for code-execution intent.
  \item Memory-poisoning and stored-prompt attacks where the injection lands in an upstream cache, RAG store, or session memory and fires on a subsequent unrelated request. The current evaluation scores each prompt independently; cross-turn or cross-session attacks are not represented.
  \item Adaptive (white-box) attackers who optimise prompts against a frozen Gate snapshot. The published benchmarks here are mostly static; competitive evaluation against an active attacker who queries the detector before crafting their payload is left for future work.
  \item Drift across model generations: the cascade is anchored to a specific reasoning-model checkpoint. Behaviour against attack distributions targeted at newer or different base models has not been measured.
  \item Prompts in domain dialects (legal contracts, clinical notes, financial filings) where lexical priors differ sharply from the general-web text used to train the public benchmarks. The evaluation under-samples these distributions.
  \item Sibling-chunk and near-duplicate leakage protection is a diagnostic, not a guarantee. Headline F1/FPR come from the StratifiedKFold pass; the parallel StratifiedGroupKFold pass is reported only as a leakage diagnostic via the F1 gap (strat - sgk). Where the composite group-key degenerates to row-identity (no chunked rows and no near-duplicate clusters) the diagnostic delta is uninformative.
\end{itemize}

\emph{Scope of the claim.} The per-dataset numbers in this paper apply to the model snapshot identified on the title page, evaluated on the public benchmark cohort listed in \S3. The deployed model is fine-tuned on additional internal data distributions that are not part of any benchmark above; production deployments may therefore report different per-dataset numbers than this paper. Where benchmarks are revised (dataset hardening, new versions) or the model is retrained, we re-evaluate and publish a revised version of this paper rather than retroactively edit existing figures.

\FloatBarrier
\clearpage
\twocolumn[\begin{@twocolumnfalse}%
\section{Results}
\begin{center}
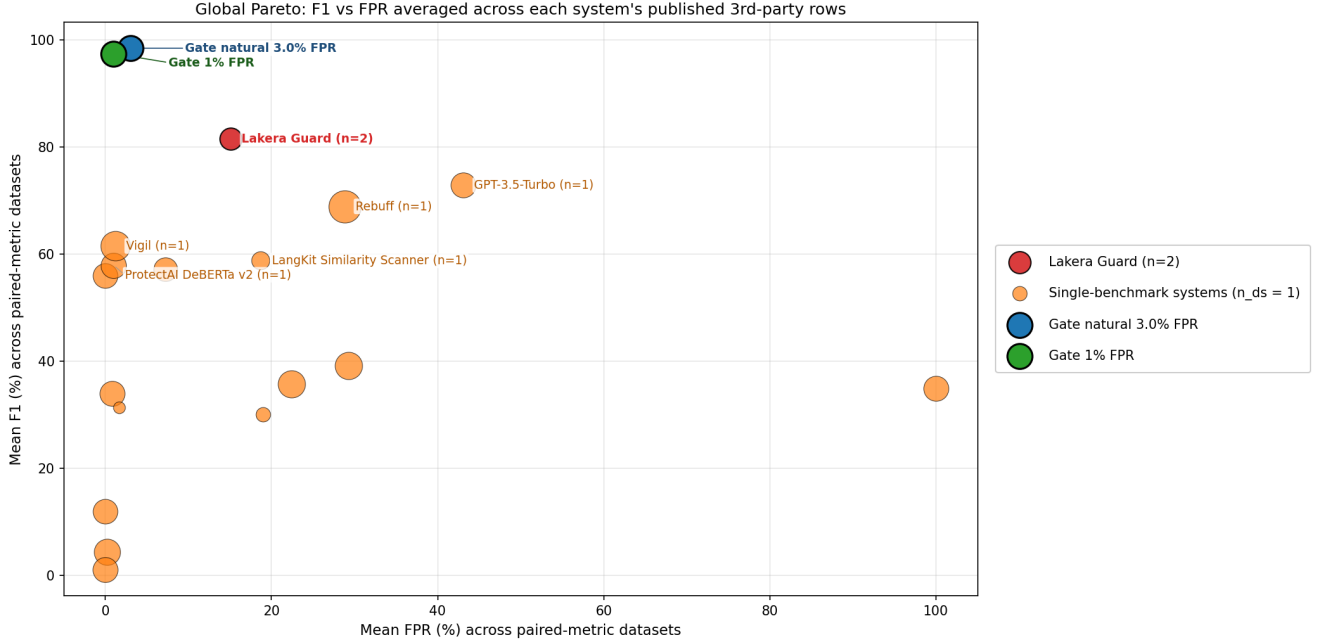
%
\paretoglobalimg%
\captionof{figure}{Global Pareto: F1 vs FPR averaged across each system's independently verified rows on the benchmarks evaluated here. Marker colours: blue $=$ Gate at the natural threshold ($\theta = 0.5$), green $=$ Gate at the 1\% FPR headline operating point, red $=$ Lakera Guard, grey $=$ other competitors with evidence on $n_{\mathrm{ds}} \geq 2$ benchmarks, orange $=$ single-benchmark competitors ($n_{\mathrm{ds}} = 1$). We keep entries where F1 and FPR are published on the same row and the source is third-party; vendor blogs and self-reported numbers are excluded. Per-system rows are averaged within a benchmark first, then across benchmarks, so each benchmark contributes once. Marker size scales with mean published competitor latency; Gate's marker uses the median competitor latency as a presentational fallback (the cascade has no single representative latency, and \S5 reports Gate's actual distribution).}%
\label{fig:pareto-global}%
\end{center}%
\vspace{0.6em}%
\end{@twocolumnfalse}]

Two operating points anchor the comparison. The \emph{headline}
threshold $\theta_{1\%}$ is the value picked on
the held-out folds that bounds pooled FPR at $\le 1\%$,
applied uniformly to every dataset. The \emph{natural} threshold
$\theta_{F_1}$ is the unconstrained F1-maximiser ($\theta = 0.5$
on the calibrated probability). The table below reports both,
for both micro (pooled confusion matrix) and macro (per-source
F1 then averaged) aggregations across the 16 public
datasets (12,111 samples, 5-fold StratifiedKFold (by row)).

\begin{center}
\resizebox{\linewidth}{!}{%
\begin{tabular}{l rrrr rrr}
\toprule
 & \multicolumn{4}{c}{Micro (pooled)} & \multicolumn{3}{c}{Macro (mean across sources)} \\
\cmidrule(lr){2-5}\cmidrule(lr){6-8}
Operating point & $F_1$ & P & R & FPR & $F_1$ & R & FPR \\
\midrule
Headline (FPR $\le 1\%$) & 97.4\% & 99.6\% & 95.4\% & 1.0\% & 95.7\% & 92.7\% & 0.9\% \\
Natural ($\theta = 0.5$, max-$F_1$) & 98.7\% & 98.5\% & 98.9\% & 4.2\% & 98.4\% & 98.3\% & 4.4\% \\
\bottomrule
\end{tabular}
}
\end{center}

\noindent At the headline operating point, Gate ranks \#1 on
8 ranked datasets, \#2 on 3, \#3 on 1; 1 dataset (hackaprompt) has no published third-party comparator with a comparable metric and is omitted from the per-dataset table.

\subsection{Per-dataset leaderboard}
\label{sec:leaderboard}

Each dataset uses its \emph{primary metric} (F1 for balanced
splits, Recall for all-attack collections, Over-Defense
Accuracy for all-benign sets; asterisk = self-evaluation, treat
as upper bound). The per-system per-dataset Pareto below plots
one point per $(\text{system} \times \text{dataset})$ pair;
the aggregated mean Pareto at the top of the section is the same
data projected onto a single point per system. The per-dataset
table grounding these plots lives in Section~3 (Data).


\IfFileExists{plots/brainstorm-det-loglog.png}{%
\begin{figure}[H]
  \centering
  \includegraphics[width=\linewidth,height=0.32\textheight,keepaspectratio]{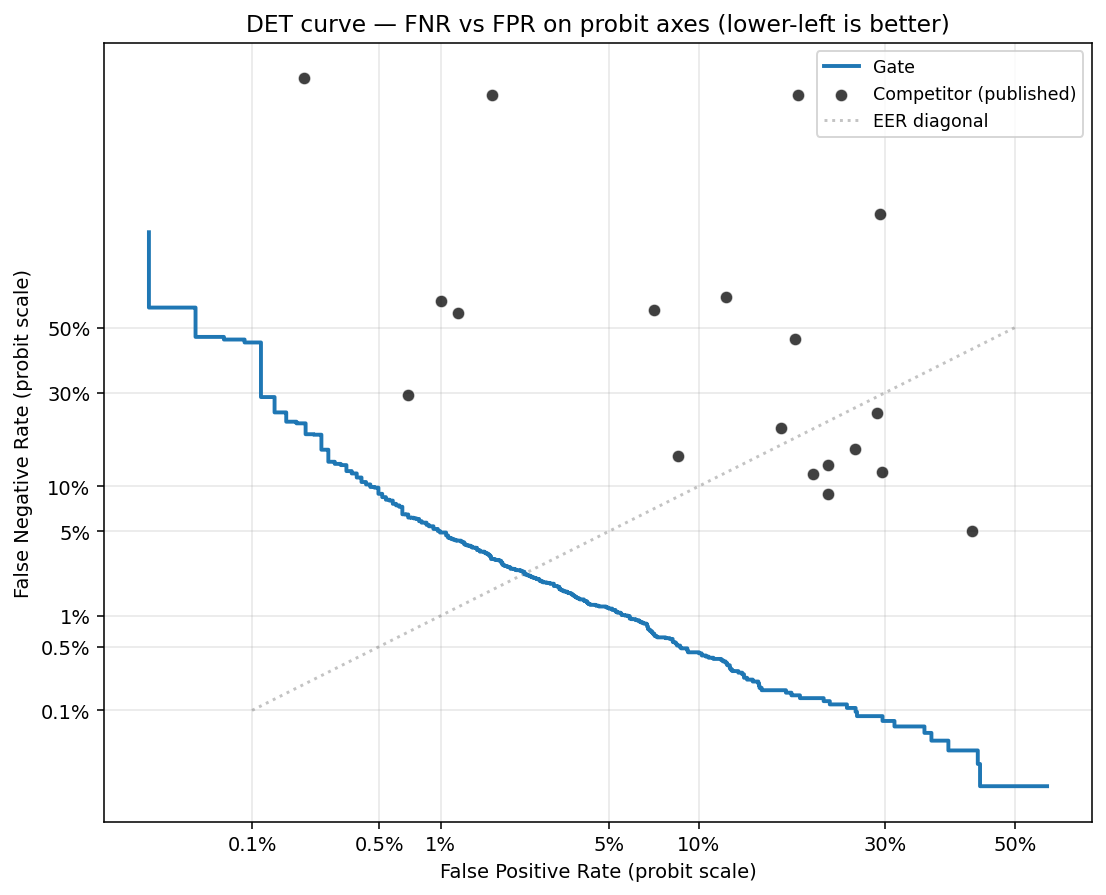}
  \caption{Detection Error Tradeoff on probit axes. Probit scaling stretches the low-FPR regime where competitor differentiation happens.}
  \label{fig:det-loglog}
\end{figure}
}{}
\IfFileExists{plots/brainstorm-tpr-at-fpr-budget.png}{%
\begin{figure*}[!t]
  \centering
  \includegraphics[width=0.85\linewidth,height=0.30\textheight,keepaspectratio]{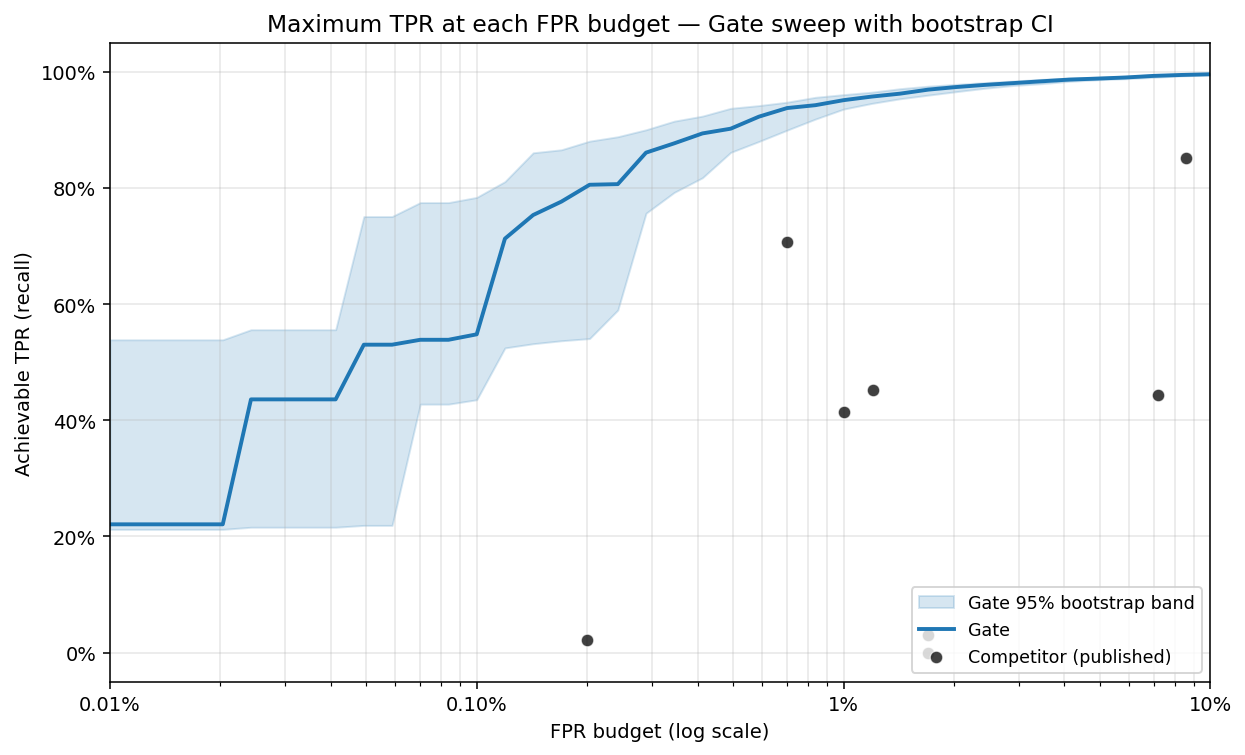}
  \caption{Maximum TPR at each FPR budget. Gate sweeps $\theta$ along the curve with a bootstrap band; black dots are individual competitors.}
\end{figure*}
}{}
\IfFileExists{plots/all-datasets-pareto.png}{%
\begin{figure*}[!t]
  \centering
  \includegraphics[width=0.85\linewidth,height=0.36\textheight,keepaspectratio]{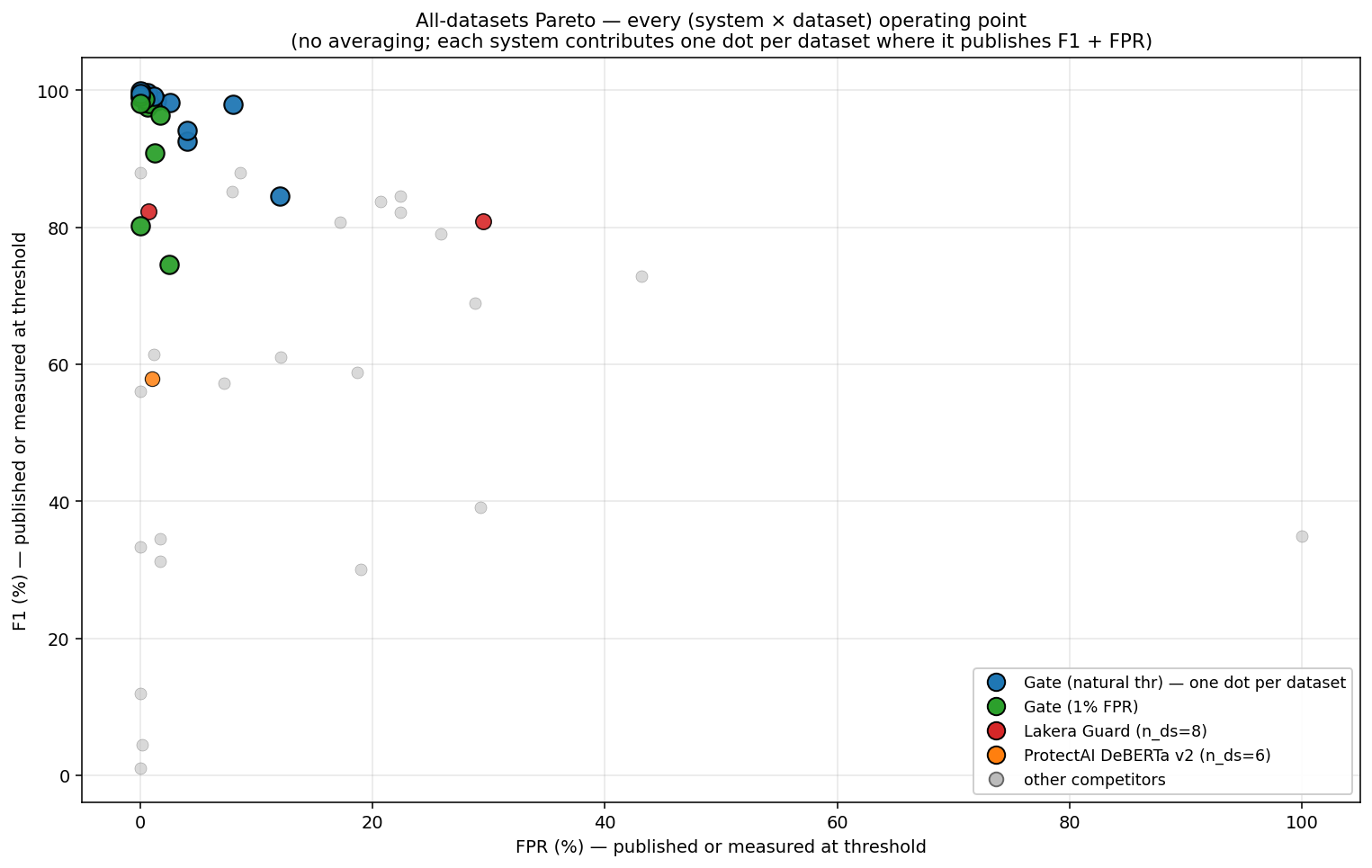}
  \caption{Every (system $\times$ dataset) operating point. Gate at natural and at FPR-$1\%$; Lakera Guard in red; second-most-published competitor in orange; everything else grey.}
  \label{fig:pareto-per-dataset}
\end{figure*}
}{}

\FloatBarrier
\subsection*{Head-to-head: Lakera Guard}

Lakera Guard is the most-cited commercial competitor in this space and publishes per-dataset numbers on the broadest set of benchmarks. The figures below report the head-to-head comparison at \emph{per-dataset matched FPR} so the comparison is like-for-like regardless of where Lakera placed its operating point. Two caveats apply on a subset of rows. (a) When the dataset's primary metric is FPR or over-defense-accuracy (notinject, wildguard-benign), matching FPR collapses the comparison, so Gate's value at the global headline threshold is shown instead, marked $^\dagger$. (b) When Lakera published the primary metric but no FPR for a dataset (the three gentel-bench splits), Gate's threshold is matched to a 1\% FPR surrogate rather than Lakera's unknown operating point, marked $^\ddagger$. The footnote under each figure enumerates which rows are true matched-FPR comparisons.

\paragraph{A note on Lakera's marketed operating point.} Lakera's product literature\footnote{\url{https://www.lakera.ai/blog/lakera-guard-fall-25-adaptive-at-scale}} advertises an FPR as low as 0.1--0.2\%. That figure comes from Lakera's internal \emph{Challenging Moderation Benchmark}, with the model's adaptive-calibration tuned per customer as part of onboarding; it is neither an out-of-the-box number nor a third-party measurement on a public benchmark. The matched-FPR figures below evaluate Lakera at its third-party published FPR on each public benchmark (arXiv:2505.13028 Table 5~\cite{c1} on \texttt{deepset}, arXiv:2409.19521 Table 2 (Goal Hijacking Attack)~\cite{c18} on \texttt{gentel-goal-hijacking}, arXiv:2409.19521 Table 1 (Jailbreak Attack)~\cite{c18} on \texttt{gentel-jailbreaking}, arXiv:2409.19521 Table 3 (Prompt Leaking Attack)~\cite{c18} on \texttt{gentel-prompt-leaking}, arXiv:2603.13247 Table 3 (ILION-Bench v2 comparative evaluation)~\cite{c3} on \texttt{ilion-bench}, arXiv:2410.22770 Table 1~\cite{c11} on \texttt{notinject}, and arXiv:2410.22770 Table 7~\cite{c11} on \texttt{wildguard-benign}). We include the note for transparency; the figures below report what this evaluation framework can measure directly: third-party numbers on public benchmarks at a matched operating point.

\needspace{8\baselineskip}
\subsection*{Sample-size-matched comparisons}

Comparing a frozen published score on $n_\text{theirs}$ samples to our number on a larger trace is not strictly apples-to-apples; a smaller evaluation set carries wider variance. For every competitor that published a smaller $n$ than ours on the same dataset, we draw $B$ subsamples of size $n_\text{theirs}$ from our out-of-fold predictions and bracket the comparison with a 95\% percentile interval. The reader can compare the competitor's point estimate against the matched-$n$ interval directly: when the published value sits inside the interval the comparison is within sampling noise.

\paragraph{ilion-bench.} Primary metric: f1; our trace: $n = 400$.

{\footnotesize\setlength{\tabcolsep}{3pt}\begin{tabular*}{\linewidth}{@{\extracolsep{\fill}}p{0.30\linewidth}rrcr@{}}
\toprule
Comp. & $n_\text{t}$ & Theirs & Matched-$n$ 95\% CI & Full-$n$ \\
\midrule
Lakera Guard & 380 & 0.809 & $[0.837, 0.854]$ & 0.845 \\
OpenAI Moderation API & 380 & 0.119 & $[0.837, 0.855]$ & 0.845 \\
Llama Guard 3 & 380 & 0.011 & $[0.837, 0.855]$ & 0.845 \\
ILION & 380 & 0.852 & $[0.837, 0.855]$ & 0.845 \\
\bottomrule
\end{tabular*}}

\begin{figure}[H]
\centering
\includegraphics[width=\linewidth]{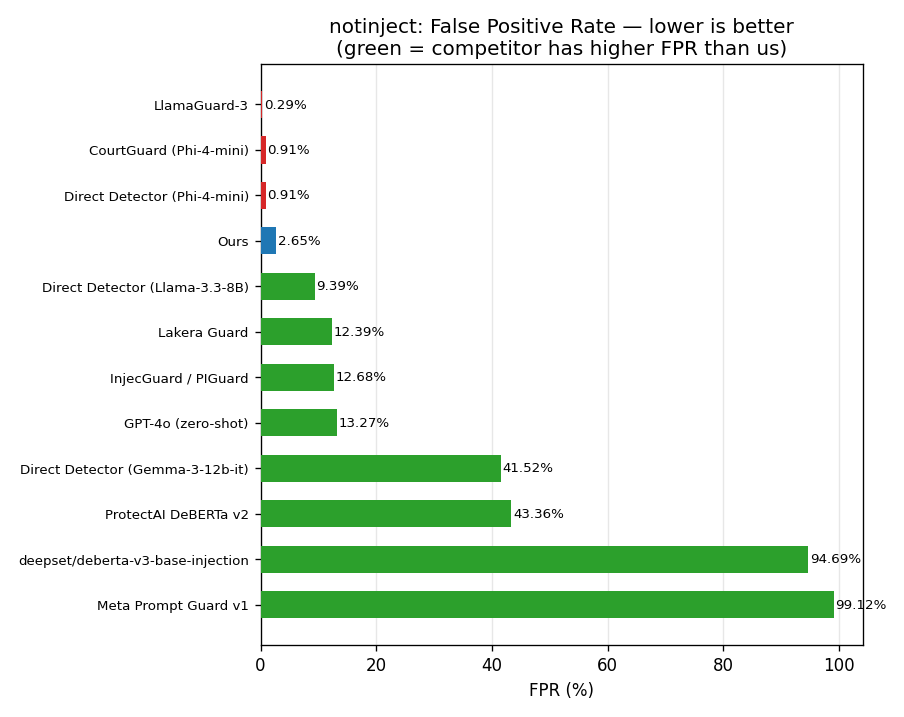}
\caption{NotInject: false-positive rate against every published competitor. Lower is better. Blue bar is the system under test at the global headline threshold; green bars are competitors with higher FPR than us, red bars lower.}
\end{figure}

\begin{figure}[H]
\centering
\includegraphics[width=\linewidth]{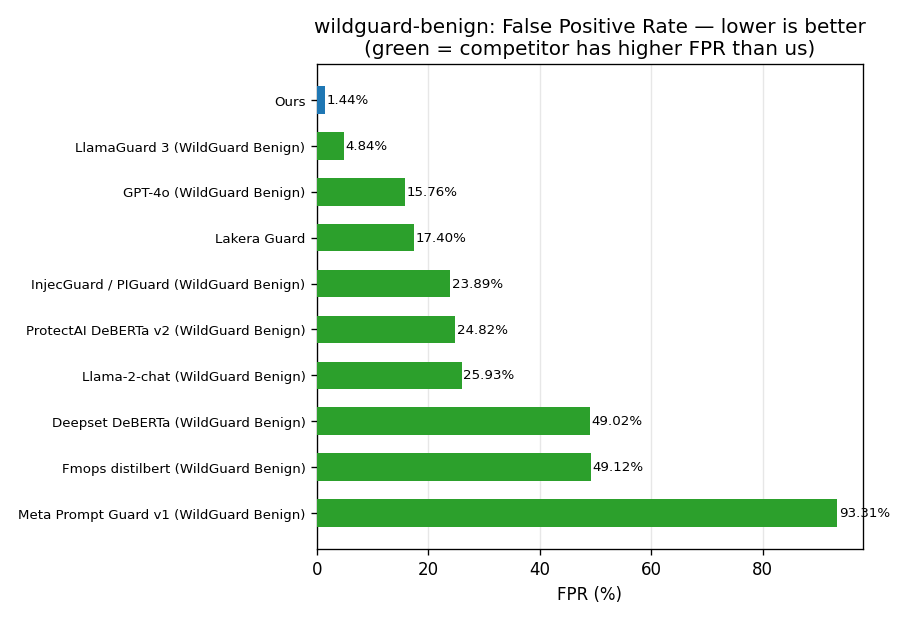}
\caption{WildGuard-benign: false-positive rate against every published competitor. Same colour rules as the NotInject chart above.}
\end{figure}

\begin{figure*}[!tb]
\centering
\includegraphics[width=\textwidth,height=0.45\textheight,keepaspectratio]{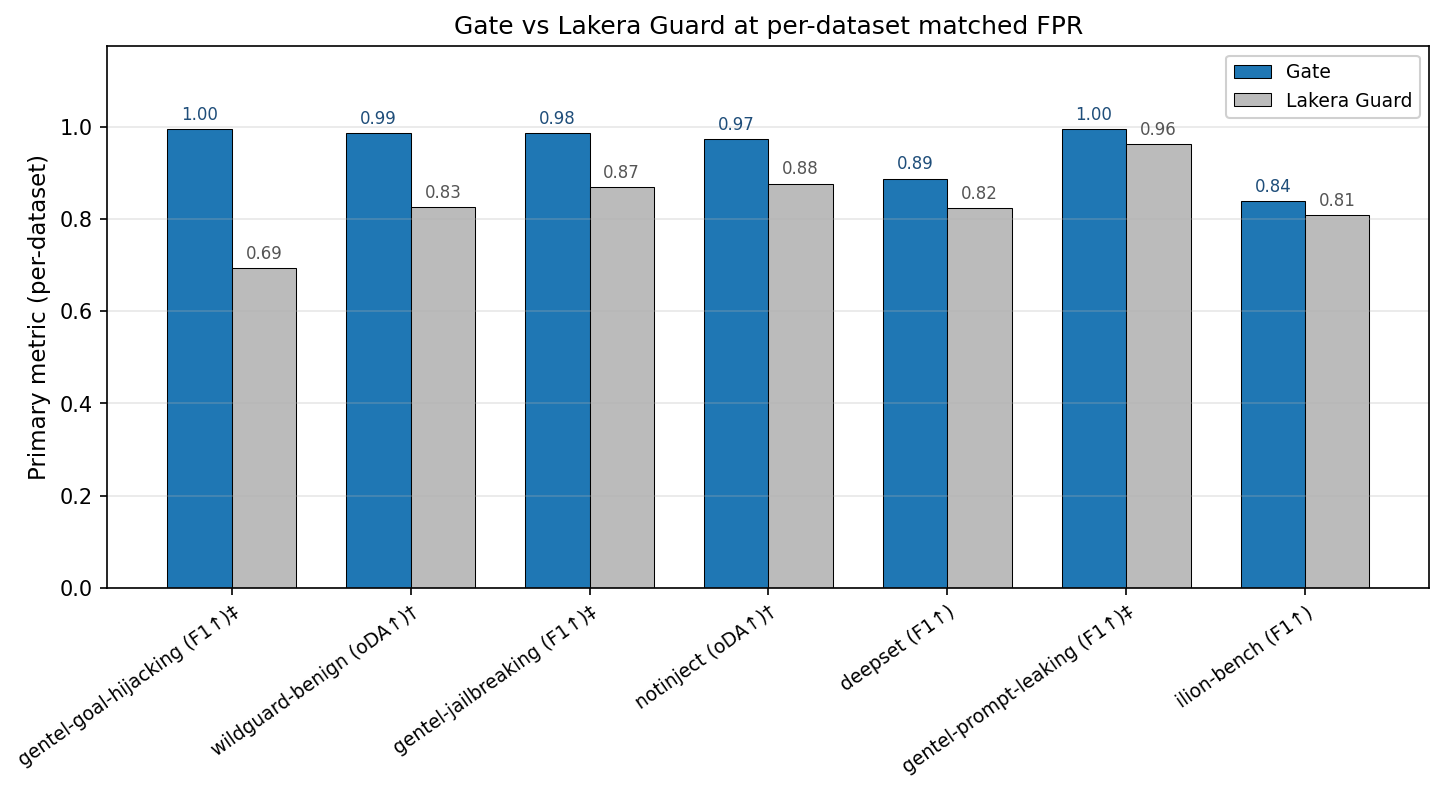}
\caption{Gate vs Lakera Guard: per-dataset values at matched FPR. Bars are the per-dataset primary metric (F1 for balanced splits, over-defense accuracy for all-benign splits) with Lakera's published FPR used as the threshold target. Rows marked $^\dagger$ have FPR / over-defense-accuracy as the primary metric, where matching FPR collapses the comparison, so Gate's value at the global headline threshold is shown instead. Rows marked $^\ddagger$ are gentel-bench splits where Lakera published the primary metric but no FPR, so Gate's threshold is matched to a 1\% FPR surrogate rather than to Lakera's unknown operating point. Unmarked rows are true matched-FPR comparisons; the per-dataset table on the preceding page enumerates which rows fall into each bucket.}
\end{figure*}

\paragraph{What the figure shows.} At matched FPR, Gate is at or above Lakera Guard on every unmarked row in the figure --- the bars give the head-to-head magnitude per dataset. The largest improvements appear on the multi-attack benchmarks where Lakera's headline operating point is tuned for a different FPR than ours; matching at Lakera's published FPR moves both points onto a like-for-like axis. Dagger rows are excluded from the headline matched-FPR tally because the primary metric is FPR or over-defense-accuracy (the comparison collapses when FPR itself is matched), and double-dagger rows are gentel-bench splits where Lakera published the primary metric but no corresponding FPR.

\paragraph{Caveats.} The head-to-head is bounded by what each side publishes. Lakera's per-dataset numbers come from the third-party measurements cited above; Gate is re-evaluated against those frozen reference points on the same public traces, so the comparison is point-in-time and tied to the Lakera version those papers measured. The figure isolates detection at a matched operating point --- latency, throughput, and serving cost are reported separately in \S5 and are not part of this matched-FPR claim. Benchmarks where Lakera does not publish are absent from the figure rather than reported as wins.

\FloatBarrier

\FloatBarrier

\FloatBarrier

\FloatBarrier

\FloatBarrier
\needspace{8\baselineskip}
\section{Latency}

End-to-end latency is measured under the deployed routing policy: per-stage samples come from concurrency sweeps on the production hardware, and the per-request composite is drawn under the cascade's actual exit policy. The distribution is bimodal --- a fast-mode cluster ($\approx$90\% of requests, first stage only) and a deferred-mode tail ($\approx$10\%) that incurs a second per-request analysis step. Numbers are saturation-free percentiles on warm pods; cold-start, bursty-queue contention, and cross-AZ jitter are not modelled and would shift the tail above the table.

\begin{tabular}{lr}
\toprule
Distribution & End-to-end latency \\
\midrule
Median (p50) & \textbf{53\,ms} \\
p90 & 60\,ms \\
p95 & \textbf{571\,ms} \\
p99 & 612\,ms \\
p99.9 & 685\,ms \\
Mean & \textbf{104\,ms} \\
Max (observed) & 699\,ms \\
\bottomrule
\end{tabular}

\IfFileExists{plots/cascade-histogram.png}{%
\begin{figure}[H]
  \centering
  \includegraphics[width=\linewidth]{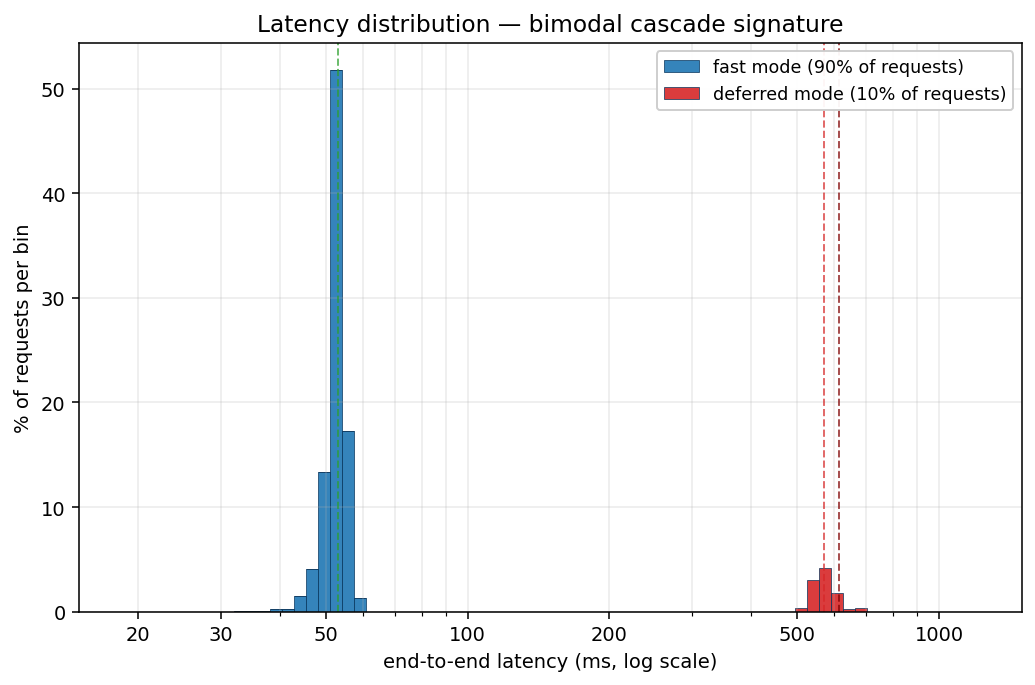}
  \caption{Latency density distribution: density histogram of end-to-end latency (log x-axis, linear y-axis).}
  \label{fig:cascade-histogram}
\end{figure}
}{}

Against published competitor latencies, Gate's mean of 104\,ms is the lowest in the corpus. Each competitor figure is the arithmetic mean of every per-dataset latency we have in the cited papers; the parenthesised range shows the min--max spread. Closest: Lakera Guard (mean 140\,ms across $n=6$ per-dataset measurements, range 51--305\,ms). Slowest: Rebuff (mean 23898\,ms across $n=2$ per-dataset measurements, range 18465--29330\,ms). Competitor papers do not label percentiles (likely mean-of-N on the cited hardware).

\IfFileExists{plots/latency-vs-competitors.png}{%
\begin{figure*}[!t]
  \centering
  \includegraphics[width=\linewidth,height=0.42\textheight,keepaspectratio]{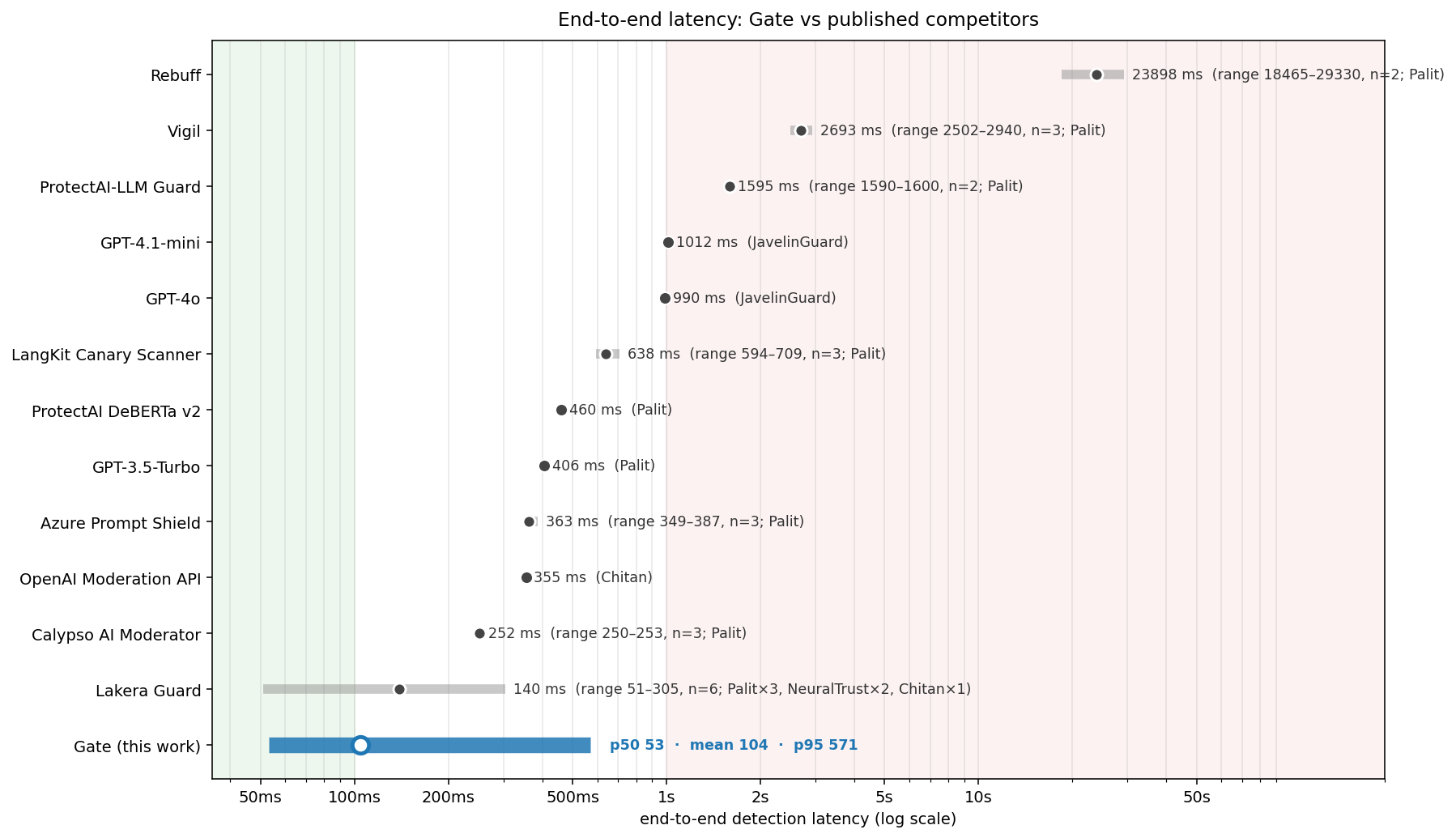}
  \caption{End-to-end detection latency: Gate (blue, p50$\to$p95 with mean marker) vs published competitors (grey min--max range or single dot). Log x-axis. Sources: Palit et al.~\protect\cite{c1} (Tables~3, 5), JavelinGuard~\protect\cite{c2} (Table~4), ILION-Bench v2 / Chitan~\protect\cite{c3} (Table~3 --- OpenAI Moderation and Lakera baselines reported by an author of a competing system; flagged 3rd-party but the surrounding context is vendor-biased), and NeuralTrust~\protect\cite{c4} firewall comparison. ``$\times N$'' counts data points per source.}
  \label{fig:latency-vs-competitors}
\end{figure*}
}{}

\FloatBarrier

\FloatBarrier
\section{Conclusion}

Two design choices drive the headline numbers reported here. The
first is treating prompt-injection detection as a \emph{cascade
ensemble} problem rather than a single-model classification
problem: signals from independent classifier heads, an embedding
representation, and a deferred second-stage analysis path are
combined under a global decision rule, so each component
contributes within its strength and the system as a whole inherits
none of any single component's blind spots. The second is enforcing
a single global operating point during evaluation: one threshold
applied uniformly across every benchmark, selected on held-out
out-of-fold predictions subject to a fixed FPR cap, rather than
per-dataset tuning. The combination yields a micro-F1 of
\textbf{97.4\%} at the deployable headline
operating point (FPR $\le 1\%$, pooled
1.0\%), and
\textbf{98.7\%} at the unconstrained
$F_1$-maximising natural threshold ($\theta=0.5$, pooled
4.2\%), across 12,111 samples from
16 public benchmarks. The system ranks \#1 on
8 ranked benchmarks and within 95\% bootstrap noise on most of
the rest.

The methodology checks support those numbers as well as
the model. Headline metrics are reported with stratified bootstrap
95\% confidence intervals; a parallel StratifiedGroupKFold pass
with a composite (parent-prompt id + MinHash near-duplicate) key
runs alongside the headline StratifiedKFold pass so the gap
quantifies any residual leakage; random-label, adversarial-validation,
length-bias, threshold-transferability, train-vs-OOF agreement,
and prevalence-ratio diagnostics each carry a precise pass
criterion and run on every release; per-fold threshold stability
is reported numerically so the operating point is auditable;
external comparisons are evaluated at \emph{per-dataset matched
FPR} where competitors publish an FPR, and as small-$n$ matched
bootstrap intervals where they don't, so head-to-head wins are
bracketed by noise rather than asserted point-estimate-to-point-estimate.
End-to-end latency is \textbf{53\,ms p50, 104\,ms mean, 571\,ms p95} at the deployed operating point.

The biggest unmodelled risk is pretraining contamination of the
benchmark corpora themselves; the largest in-scope follow-up is
held-out evaluation on attack patterns generated after
foundation-model training cutoffs.

\clearpage
\onecolumn
\appendix
\section{Glossary}
\label{sec:glossary}

\begin{description}[leftmargin=2.0em,labelindent=0pt,style=nextline,itemsep=0.25em]
\item[F1] harmonic mean of precision and recall: $F_1 = 2 \cdot P \cdot R / (P + R)$.
\item[Precision (P)] fraction of flagged prompts that were actually injections: $P = \mathrm{TP} / (\mathrm{TP} + \mathrm{FP})$.
\item[Recall (R)] fraction of injections caught: $R = \mathrm{TP} / (\mathrm{TP} + \mathrm{FN})$.
\item[FPR] false-positive rate: $\mathrm{FPR} = \mathrm{FP} / (\mathrm{FP} + \mathrm{TN})$.
\item[AUC] area under the ROC curve. 0.5 is chance, 1.0 is perfect.
\item[Over-Defense Accuracy (ODA)] for all-benign datasets, $\mathrm{ODA} = 1 - \mathrm{FPR}$.
\item[$K$-fold CV] partition the trace into $K$ disjoint folds; hold each out for evaluation while training on the rest.
\item[StratifiedKFold] $K$-fold variant that preserves the label marginal $P(y \mid \mathcal{D}_k) \approx P(y \mid \mathcal{D})$; the headline pass that feeds the abstract and leaderboard.
\item[StratifiedGroupKFold] $K$-fold variant that preserves the label marginal \emph{and} enforces a group constraint so members of one group (parent prompt + near-duplicate cluster) cannot straddle folds; reported as a leakage diagnostic via $\Delta F_1 = F_1^{\text{strat}} - F_1^{\text{sgk}}$.
\item[OOF prediction] out-of-fold prediction: for each row, the prediction made by a model trained on the $K-1$ folds that excluded it.
\item[Stratified bootstrap CI] confidence interval from $B$ resamples drawn within stratification cells (here: source $\times$ label); percentile interval at $2.5\%$ / $97.5\%$.
\item[LODO] leave-one-dataset-out cross-validation. Train on every other source; evaluate on the held-out source.
\item[ECE] Expected Calibration Error: weighted average of $|\operatorname{acc}(B_m) - \operatorname{conf}(B_m)|$ across confidence bins.
\item[Brier score] mean squared error between predicted probabilities and binary labels: $\frac{1}{N}\sum_i (\hat p_i - y_i)^2$.
\item[Operating point] the threshold (or threshold combination) at which the system commits to hard labels. Chosen here at a fixed FPR target on out-of-fold predictions, applied uniformly across every dataset.
\item[Self-evaluation (*)] a competitor score reported by the system's own authors; treated as an upper bound.
\end{description}

\end{document}